\documentclass[acmtog,nonacm]{acmart}

\AtBeginDocument{%
  \providecommand\BibTeX{{%
    \normalfont B\kern-0.5em{\scshape i\kern-0.25em b}\kern-0.8em\TeX}}}

\usepackage{amsmath,amsfonts}
\usepackage{algorithm}
\usepackage{array}
\usepackage[caption=false,font=normalsize,labelfont=sf,textfont=sf]{subfig}

\usepackage{tabularx}

\usepackage{textcomp}
\usepackage{stfloats}
\usepackage{url}
\usepackage{verbatim}
\usepackage{graphicx}
\usepackage{hyperref}
\usepackage{amsmath}
\usepackage{bm}
\usepackage{subfig}
\usepackage{enumitem,kantlipsum}
\UseRawInputEncoding
\usepackage{multirow}
\usepackage{colortbl}

\definecolor{lightBlue}{rgb}{.62109375,.75390625,.82421875}
\definecolor{lightPink}{rgb}{.914,.804,.816}

\definecolor{greendatasets}{rgb}{.914,.804,.816}

\usepackage{pifont}
\newcommand{\cmark}{\ding{51}}%
\newcommand{\xmark}{\ding{55}}%

\begin{document}

\title{CNN-Based Action Recognition and Pose Estimation for Classifying Animal Behavior from Videos: A Survey}


\author{Michael P\'{e}rez}
\orcid{0000-0002-2453-0933}
\author{Corey Toler-Franklin}
\orcid{0000-0002-8240-3756}
\email{ctoler@cise.ufl.edu}
\affiliation{%
\institution{University of Florida}
\country{USA}}

\renewcommand{\shortauthors}{Perez, M.  et al.}

\begin{abstract} Classifying the behavior of humans or animals from videos is important in biomedical fields for understanding brain function and response to stimuli. Action recognition, classifying activities performed by one or more subjects in a trimmed video, forms the basis of many of these techniques. Deep learning models for human action recognition have progressed significantly over the last decade. Recently, there is an increased interest in research that incorporates deep learning-based action recognition for animal behavior classification. However, human action recognition methods are more developed. This survey presents an overview of human action recognition and pose estimation methods that are based on convolutional neural network (CNN) architectures and have been adapted for animal behavior classification in neuroscience. Pose estimation, estimating joint positions from an image frame, is included because it is often applied  before classifying animal behavior. First, we provide foundational information on algorithms that learn spatiotemporal features through \hbox{2D}, two-stream, and \hbox{3D} CNNs. We explore motivating factors that determine optimizers, loss functions and training procedures, and compare their performance on benchmark datasets. Next, we review animal behavior frameworks that use or build upon these methods, organized by the level of supervision they require. Our discussion is uniquely focused on the technical evolution of the underlying CNN models and their architectural adaptations (which we illustrate), rather than their usability in a neuroscience lab. We conclude by discussing open research problems, and possible research directions. Our survey is designed to be a resource for researchers developing fully unsupervised animal behavior classification systems of which there are only a few examples in the literature.  
\end{abstract}

\ccsdesc[500]{Computing methodologies~Neural networks}

\keywords {Deep Learning, Action Recognition, Pose Estimation, Behavior Phenotyping, Levels of Supervision}

\maketitle

\section{Introduction} \label{intro}

\begin{figure}[t]
\begin{center}
\includegraphics[width=\linewidth]{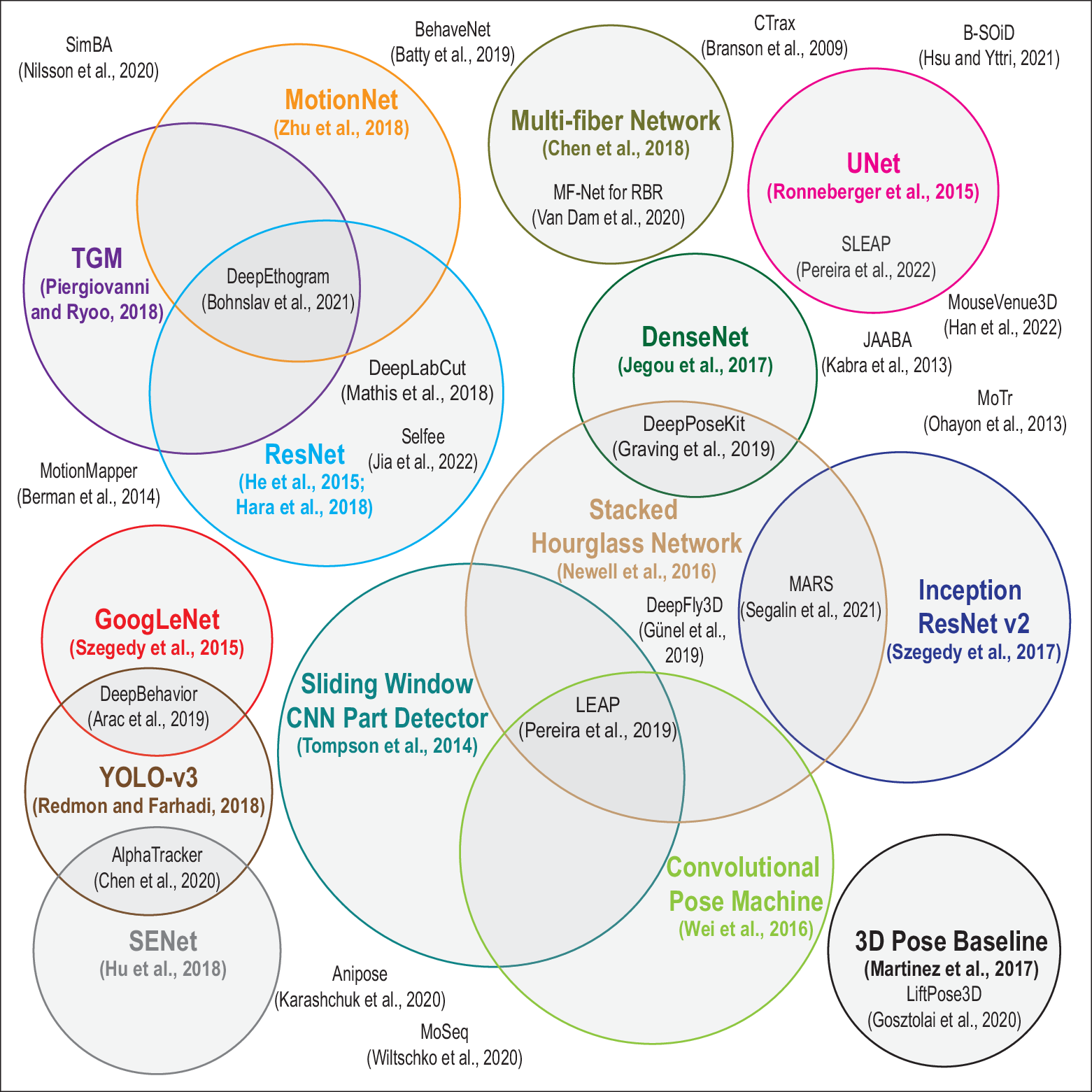}
\caption{CNN-based Animal Behavior Classification Frameworks and Animal Pose Estimation and Tracking Methods Organized by the Neural Network Architectures They Build Upon. Representative Examples Shown.} 
\label{fig:venndiagram}
\end{center}
\end{figure}

This survey presents a comprehensive overview of CNN architectures for human action recognition and human pose estimation from videos. The paper traces the development and adaptations of the subset of networks that have been extended for animal behavior classification. Action recognition aims to identify and classify the activity in a trimmed video. The output of an action recognition algorithm is a set of class labels for each trimmed video. When the class label is one of a set of pre-defined expressions that occur in response to a stimuli (behavior) ~\cite{coria2022neurobiology}, the task is behavior classification. Pose estimation is the task of determining a subject's position and orientation in an image frame. Pose-based action recognition methods use pose estimation predictions (image frame coordinates) to classify body movements. Pose estimation may be used to identify joint positions before classifying behavior.  Action recognition is critical for analyzing motion in videos for disciplines such as robotics, security, and biomedical engineering. These methods use labels. Because manual labeling is time consuming, and introduces human bias and reproducibility issues, there is a growing interest in developing action recognition methods for fully unsupervised behavior classification that have less complex systems. A fully unsupervised method would not require labels, hand-crafted features, or additional pose estimation steps. This survey reviews CNN-based human action recognition and pose estimation methods and their extensions to animal behavior classification with a particular focus on open problems with unsupervised animal behavior classification in neuroscience. 

\vspace{10pt}
\noindent\textbf{Our contributions include:} 

\begin{enumerate}

\item A review of \hbox{2D}, two stream and \hbox{3D} CNNs that detect and analyze spatio-temporal features in videos for action recognition and pose estimation (Sections~\ref{sec:hpe} and~\ref{sec:actionrecognition}), including open challenges, design decisions, key contributions, and performance on benchmark datasets (Tables~\ref{tab:pose-estimation-summary-table} and~\ref{tab:action-recognition-summary-table}). The review focuses on the subset of methods that have extensions for classifying animal behavior.  

\item An organizational strategy for categorizing a representative  set of animal behavior classification frameworks by the level of supervision they require, considering their dependency on handcrafted features, labels, pose estimation and learning strategies (Section~\ref{sec:taxonomy}). Key contributions and performance on benchmark datasets are summarized (Tables~\ref{tab:poseaccuracy},~\ref{tab:table-animal-pose-estimation}, and~\ref{tab:table-animal-action}).

\item Illustrations that permit visual comparison of animal behavior classification frameworks, including system components like pose estimation techniques, neural network architectures and dimensionality reduction and clustering algorithms (Figures~\ref{fig:posebased},~\ref{fig:SNoPE}~\ref{fig:USNoPE}). Diagrams that show commonalities between underlying CNN architectures (Figures~\ref{fig:venndiagram},~\ref{fig:actionrecognition}), and  relationships with the human action recognition and pose estimation methods they build upon (Figure~\ref{fig:animalbehaviormodelaction}).

\item A discussion of opportunities to develop or extend action recognition methods motivated by open problems with unsupervised animal behavior classification. 

\end{enumerate}

\subsection{Motivation}\label{sec:motivation}

Action recognition is an important research area in computer vision  that continues to experience significant growth. This is due in part to its broad application in fields such as robotics, security, and biomedical engineering for tasks such as human-robot collaboration~\cite{Li2021d}, surveillance~\cite{Babiker17}, and behavior analysis~\cite{Segalin2021,Arac2019,coria2022neurobiology}. Large publicly available video datasets~\cite{Karpathy2014} with high inter-class variability have facilitated the development of deep learning algorithms for video classification that are more robust and effective. However, open problems remain. 

This paper focuses on action recognition methods that use CNNs to learn spatio-temporal features from videos. Spatio-temporal features are key points that exhibit spatial variations in color intensities within a frame, and temporal variation between frames. Neural networks detect these features and use them to classify video clips. Such features may be learned through network architectures that have both a spatial stream that operates on individual frames, and a temporal stream that operates on motion information. Other modifications add a temporal dimension to the filters and pooling kernels to create \hbox{3D} CNNs.

\begin{figure}[t]
\begin{center}
\includegraphics[width=1\linewidth]{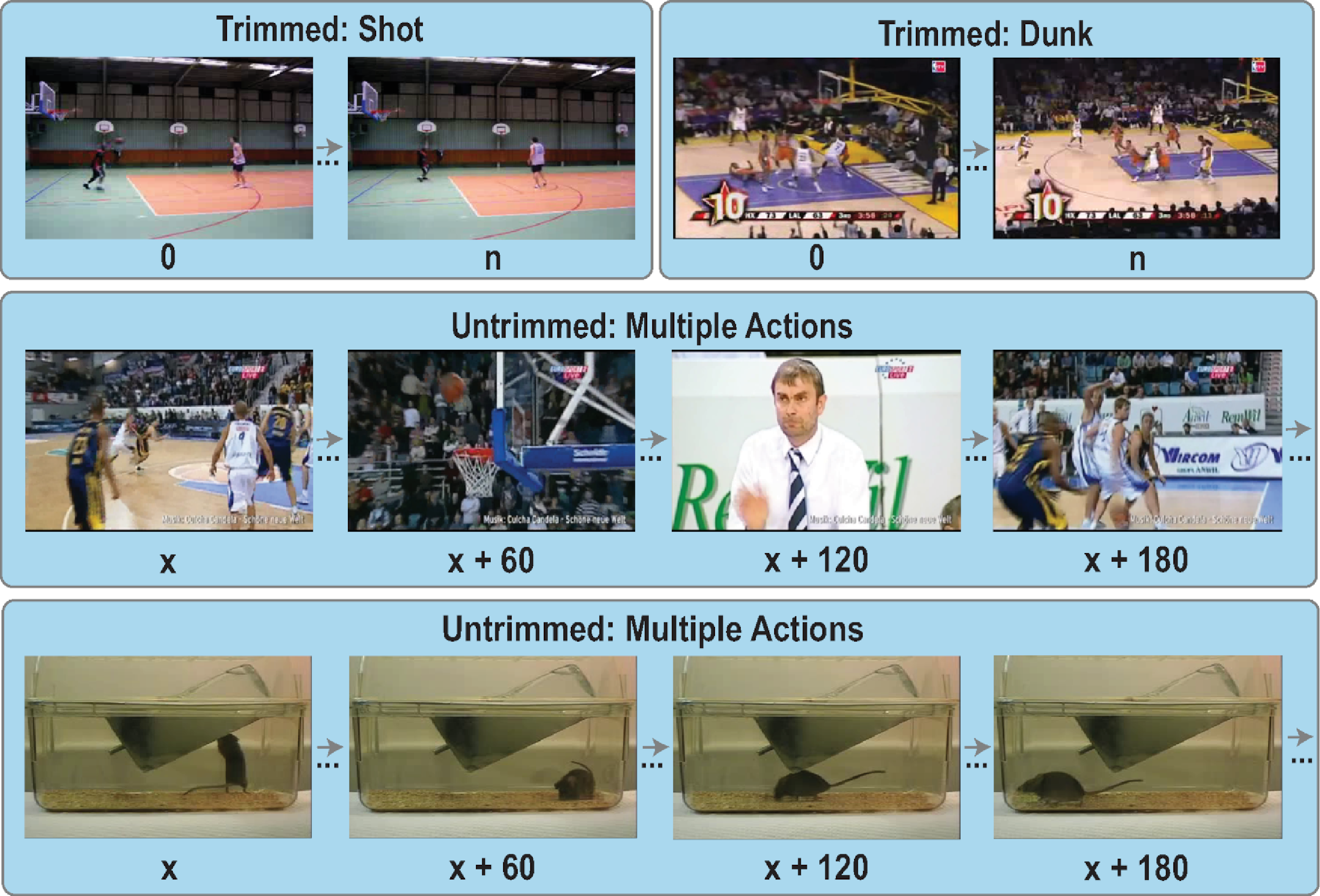}
\caption{Trimmed and Untrimmed Videos: (top row) The first and last frames from clipped videos of (left) a basketball dunk and (right) a basketball shot from UCF101~\cite{Soomro2012}. (middle row) Four frames $60$ frames apart from an untrimmed video from THUMOS14~\cite{Jiang2014}. (bottom row) Four frames $60$ frames apart from a untrimmed animal behavior dataset~\cite{Jhuang2010}.}
\label{fig:trimmedanduntrimmed}
\end{center}
\end{figure}

Action recognition algorithms operate on trimmed videos. These trimmed videos contain one action instance. Untrimmed videos are long unsegmented videos, containing multiple action instances~\cite{Vahdani2022} (Figure~\ref{fig:trimmedanduntrimmed}). Action detection (localization, or spotting) is a related research area that locates actions of interest in space and (or) time in both trimmed and untrimmed videos. Before classifying the actions in an untrimmed video, the start and end time of each action is determined. Many classification frameworks we review in Section~\ref{sec:taxonomy} operate on untrimmed videos. A small number may even incorporate networks inspired by action detection. However, they generally produce per-frame labels, without the complexities of an additional temporal action detection~\cite{Vahdani2022}  step to segment videos. For this reason, we do not include action detection in our review. DeepEthogram~\cite{Bohnslav2021} is an interesting example that demonstrates that action recognition and action detection are highly related. The pipeline uses a CNN developed for action recognition to estimate motion features, and then classifies the spatial and motion features in each frame using a network designed for action detection. The result is a per-frame set of behavior class labels.

Our review includes three types of learning models, categorized by the training data they operate on: \emph{supervised} that require labeled data, \emph{semi-supervised} that use both labeled and unlabeled data, and \emph{unsupervised} that operate on unstructured data without labels. 
Figure~\ref{fig:trimmedanduntrimmed} (top row) shows class labels for \emph{shot} and \emph{dunk}. Figure~\ref{fig:trimmedanduntrimmed} (bottom row, left to right) shows animal behavior classification labels for \emph{rear}, \emph{walk} , \emph{walk} , \emph{walk}.  The label \emph{rear} indicates that the animal stands on hind legs.

Most methods presented in this review require some level of supervision. However, using predefined labels is limiting, as micro-scale actions that are not in the labeled dataset but exist in the unstructured data, may go undetected. Direct processing on raw pixels would enable fewer processing steps, without the need for hand-crafted features to produce consistent poses for different subjects. 

There are many challenges with action recognition from videos. It is challenging to trace the motion of body parts which are occluded in intermittent frames. Artifacts from variations in lighting or varying parameters from camera  or devices (that have different imaging quality and frame rates) add additional challenges. Many pose-based systems do not generalize well to \emph{in-the-wild} conditions.

Behavior analysis is important~\cite{MATHIS20201,Anderson2014} with many real-world applications. Recent projects extend human action recognition for detecting and interpreting behavior in small animals. Additional challenges exist in this domain~\cite{Anderson2014}. Tracking multiple subjects that have identical appearance is challenging, often requiring specialized tracking devices that are distracting. Many systems incorporate several machine learning tasks such as object detection, pixelwise segmentation, and dimensionality reduction. Thus, classification systems are often complex, requiring many processing steps. We present an organizational strategy for animal behavior classification that considers system components and learning strategies in addition to inputs. 

There is a significant number of CNN-based action recognition and pose estimation methods for both human and animal activity recognition. This enables a thorough review of the subject.

\vspace{10pt}
\noindent\textbf{Relation to Other Surveys:} To our knowledge, this is the only survey with an organizational structure for examining commonalities between the neural network architectures of the related tasks of human action recognition and pose estimation (Figure~\ref{fig:venndiagram}) in the context of their extensions for animal behavior classification (Figure~\ref{fig:animalbehaviormodelaction}). Our survey presents a novel taxonomy for animal behavior classification  frameworks that is more detailed and provides a broader range of supervision levels that are not considered in other work~\cite{Berman2018}. We also include recent work with unsupervised animal behavior classification systems which are not expounded upon in other reviews. 

Surveys on human pose estimation~\cite{Chen2020a,Zheng2020} and action recognition are generally handled separately, and often cover a broad range of network architectures. Zheng et. al.~\cite{Zheng2020} provide a systematic review of deep learning-based \hbox{2D} and \hbox{3D} human pose estimation methods since $2014$, while Zhu and colleagues~\cite{Zhu2020} offer a comprehensive survey of deep learning methods for video action recognition. Other reviews on these topics target specific network architectures. Recently, transformers~\cite{Zhu2020,Li2021} have become the model of choice for processing sequential data. The following surveys cover transformers for action recognition~\cite{Zhu2020} and pose estimation~\cite{Li2021}. 

A related body of work reviews animal behavior classification models in neuroscience and computational ethology which is the study of animal behavior~\cite{Anderson2014}. These reviews cover many topics including measuring animal behavior across scales~\cite{Berman2018}, machine learning for animal activity recognition~\cite{Kleanthous2022}, advances in \hbox{3D} behavioral tracking~\cite{Marshall2022} for full pose measurements, monitoring with wearable devices~\cite{Kleanthous2022}, and behavior profiling~\cite{VonZiegler2021}. We focus on small laboratory animals in methods that do not require wearable tracking devices.

Many surveys in neuroscience focus on the usability of these methods as tools in a lab~\cite{MATHIS20201}. A $2021$ review~\cite{VonZiegler2021} evaluates behavior analysis approaches. We trace the technical development of the underlying neural network architectures. 

Other surveys have proposed a taxonomy for animal behavior classification systems. One proposal~\cite{Berman2018} uses a dynamical representation of behavior that describes how the measured postural time-series are changing.  Straightforward approaches~\cite{Jhuang2010, Vogelstein2014, Berman2014} manually create features which are fed to a supervised classifier or a clustering algorithm. The author argues that a dynamical representation should be produced naturally from pose dynamics in an unsupervised manner~\cite{Berman2018}.

\section{Background}\label{sec:background}

We begin with definitions, terms,  and background information that support the paper discussion.

\vspace{10pt}
\noindent\textbf{Neural networks} are integral to most frameworks in our review. Here we describe basic operations of a feedforward neural network. Given a function, $y = f^*(\bm{x})$ that maps an input vector $\bm{x}$ to an output label $y$, a feedforward neural network defines a mapping $\bm{y} = f(\bm{x}, \bm{\theta})$, and learns the values of $\bm{\theta}$ that yield the best approximation of $f^*$. Activation functions are computed at each node using function outputs from nodes at lower levels in an acyclic graph. Activation functions introduce non-linearities to improve performance on complex datasets. The rectified linear unit (ReLU)~\cite{Xu2015arxiv} function, $f(x) = \max(0,x)$ is the default activation in feedforward networks. During training, input observations $\bm{x}$ propagate forward through network layers, producing an output prediction $\bm{\hat{y}}$ and a scalar cost $J(\bm{\theta)}$. Backpropagation allows cost information to flow backwards through the network for use in optimization methods (like stochastic gradient descent) that minimize $J(\bm{\theta})$. Activation functions should avoid linearities, saturation and vanishing gradients.  The leaky ReLU function~\cite{Maas2013} is a modified ReLU function with a small slope for negative values that permits non-zero gradients, even when the unit is inactive. This avoids vanishing gradients that inhibit learning.

\vspace{8pt}
\noindent Optimizations can improve network efficiency, increase classification accuracy, and reduce adverse effects like overfitting. Architectural improvements also provide regularization within a network to address issues like vanishing gradients.

\vspace{8pt}
\noindent \textbf{Advantages of Deep Networks:} It has been shown empirically that, although deeper networks are harder to optimize, they require fewer parameters and generalize better than shallow networks~\cite{Goodfellow2016}.

\vspace{8pt}
\noindent \textbf{Regularization} is a technique that modifies a learning algorithm to reduce generalization error so that the model performs well on unseen data distributions. 

\vspace{8pt}
\noindent \textbf {Batch normalization} scales and centers inputs to normalize them for faster, more stable training. 
  
\vspace{8pt}
\noindent \textbf {Ensemble Methods} combine predictions of several neural networks to reduce test error at the expense of some additional computation costs.

\subsection{Network Architectures}

\vspace{8pt}
\noindent \textbf{Convolutional Neural Networks (CNNs):}~\cite{LeCun1989} use convolution (rather than matrix operations) in at least one layer. CNNs are regularized multi-layer perceptrons~\cite{Mcculloch1943} that leverage the spatial structure inherent in grid-like structures (like images) for regularization, through local filters, convolutions, and max-pooling layers. Advancements in GPU hardware have allowed CNNs to scale to networks with millions of parameters, improving state-of-the-art image classification and object recognition. AlexNet~\cite{Krizhevsky2012} was the first CNN to achieve state-of-the-art results on the ImageNet~\cite{Deng2009} dataset, a benchmark dataset used for a of variety deep learning challenges. The network is based on LeNet-5~\cite{LeCun1998}. VGGNet~\cite{Simonyan2015} improves upon AlexNet by using smaller convolutional filters ($3 \times 3$) and increased network depth to support between $16$ to $19$ layers.

\vspace{8pt}
\noindent \textbf{Inception}~\cite{Szegedy2015} was designed to improve utilization of computing resources within a network by using $1 \times 1$ convolutions and improved multi-scale processing. The process connects auxiliary networks to intermediate layers to encourage discrimination in the lower layers of a CNN. The auxiliary networks are discarded at test time. The GoogLeNet $22$-layer network is a notable example that won the $2014$ ImageNet challenge. Other examples like Inception-v3~\cite{Szegedy2016} improve upon the original Inception  architecture by factorizing $5 \times 5$ and $7 \times 7$ convolutions into multiple smaller, more efficient convolutional operations and incorporating batch normalization.

\vspace{8pt}
\noindent\textbf{Fast R-CNN} is a deep CNN designed for object recognition that extends the Region-Based Convolutional Neural Network (RCNN).

\vspace{8pt}
\noindent \textbf{Autoencoder:} A neural network designed for dimensionality reduction in which an encoder learns a compressed feature representation from unlabeled data. The decoder validates the representation by attempting to reconstruct the original data from the features.

\vspace{8pt}
\noindent \textbf{Residual Neural Networks (ResNets)}~\cite{He2016} utilize residual blocks with skip connections to jump over layers that are not updated to allow for better gradient to flow, mitigating the vanishing gradient problem. These architectural modifications permit deeper networks. The results improved upon VGGNet in the ImageNet challenge using a network with up to $152$ layers.  

\vspace{8pt}
\noindent \textbf{YOLOv3} (You Only Look Once)~\cite{Redmon2018arxiv} is a CNN architecture designed for fast object detection that has $53$ convolutional layers and uses residual blocks~\cite{He2016}. The YOLO family of models is not proposal-based and frames object detection as a regression problem that predicts bounding box coordinates and class probabilities given image pixels.

\vspace{8pt}
\noindent \textbf{Recurrent Neural Networks (RNNs):} These methods process sequential or time-series data. Cycles in the graph allow output from one layer to be fed back as input to another layer.

\vspace{8pt}
\noindent \textbf{Transformers}~\cite{Zhu2020,Li2021} are neural networks with self-attention~\cite{Vaswani2017} that operate on sequential data in parallel, without the need for recurrence (found in RNN-based models~\cite{Belagiannis2017,Luo2018}).

\subsection{General Classification and Training Methods}

\vspace{4pt}
\noindent \textbf{Temporal Random Forest Classifier} An extension of a random forest to the multi-frame case.

\vspace{8pt}
\noindent\textbf{Hand-crafted Features} are created by manual feature engineering. They are usually dataset dependent and do not generalize well. Neural networks, on the other hand, detect features automatically from unstructured data.

\vspace{8pt}
\noindent\textbf{Active Learning:} Semi-supervised classification of pose features as behavior labels employs unsupervised and supervised learning to minimize the number of behavioral labels while maximizing accuracy. Active learning approaches aim to systematically choose which samples should be labeled so that the number of samples is minimized, and the accuracy is maximized.

\vspace{8pt}
\noindent \textbf{Contrastive Learning:} A technique that increases the performance of vision tasks by \emph{contrasting} samples against each other. The goal is to learn attributes that are common between data classes, and attributes that make  data classes different. The approach  learns representations by minimizing the distance between positive data and maximizing the distance between negative data.

\vspace{8pt}
\noindent \textbf{Transfer learning} is a mechanism for training neural networks when there is insufficient data in the target domain.  Transfer learning uses knowledge gained from one domain to improve generalization in a related domain. For example, in a supervised learning context, a model may be trained to classify images of one object category, and then used to classify images of another (training on cats and dogs and testing on horses and cows for example). This is useful because visual categories often share low-level features like edges and shapes. Another transfer learning  technique trains a model on a large unlabeled dataset using an unsupervised learning algorithm, then fine-tunes the model on a labeled dataset with a supervised algorithm. This unsupervised pretraining is helpful as large, labeled datasets are hard to collect. 

\vspace{8pt}
\noindent \textbf{ImageNet Pre-Training:} ImageNet is a benchmark dataset used for a variety of deep learning challenges including transfer learning experiments. It contains $1,431,167$ images that correspond to $1,000$ classes and large networks trained on this dataset have shown state-of-the-art results in the image domain. The full ImageNet dataset contains $22,000$ classes, but in experiments and the ImageNet challenge $1,000$ high-level categories are used. Transfer learning methods~\cite{Oquab2014} that use ImageNet as a source domain have been effective for improving state-of-the-art classification results in related target domains. For example, a network trained on ImageNet was used~\cite{Oquab2014} for the PASCAL Visual Objects Classes challenge~\cite{Everingham2009}. Through transfer learning, deep CNNs pre-trained on ImageNet were also used for other tasks like scene recognition~\cite{Donahue2014} and object recognition~\cite{Girshick2014}. In Section~\ref{sec:actionrecognition}, we discuss transfer learning on video datasets.

\subsection{Action Recognition: Terms}

\vspace{8pt}
\noindent \textbf{Spatio-temporal Features} are used to analyze changes in image structure that vary with time. Spatial interest points occur at positions of significant local variation in image intensities~\cite{Laptev2003}. Spatio-temporal interest points are extensions of spatial interest points as they record significant local variation of the pixel intensity in the spatio-temporal domain of the video volume~\cite{Li2017SurveyOS}.

\subsection{Graphical Models}

\vspace{8pt}
\noindent \textbf{Hidden Markov Model (HMM):} A graphical model that represents probability distributions over sequences of observations that are produced by a stochastic process. The states of the process are hidden. The state $Z_t$ at time $t$ satisfies \emph{Markov properties} and depends only on the previous state, $Z_{t-1}$ at time $t-1$. 

\vspace{8pt}
\noindent \textbf{Autoregressive Hidden Markov model (AR-HMM):} A combination of autoregressive time-series and Hidden Markov Models.

\vspace{8pt}
\noindent \textbf{Graph Convolution Networks} extend convolution from images to graphs and have been used successfully in pose-based action recognition.

\vspace{8pt}
\noindent\textbf{Belief Propagation} is a message-passing algorithm for performing inference in graphical models.

\vspace{8pt}
\noindent \textbf{Graphical Models for Joint Detection:} Pose estimation methods may use graphical models with nodes that represent joints and edges that represent pairwise relationships between joints. Each final joint distribution is a product of unary terms which model the joint's appearance cues, and pairwise terms which model local joint inter-connectivites. These methods examine local image patches around a joint to detect it, and to estimate the relative position of neighboring joints.

\subsection{Pose Estimation: Terms}

\vspace{8pt}
\noindent \textbf{Part Affinity Fields} are a set of \hbox{2D} vector fields that represent the location and orientation of limbs in an image.

\subsection{Pose Estimation: Evaluation Metrics}

\noindent \textbf{Percentage of Correct Parts (PCP)} evaluates limb detection. A limb is detected if the distance between the prediction and the true limb location is within a threshold: half the limb length.

\vspace{8pt}
\noindent \textbf{Percent of Detected Joints (PDJ)} evaluates joint detection. A joint is detected if the distance between the prediction and the true joint location is within a threshold: a specified fraction of the torso diameter.

\vspace{8pt}
\noindent \textbf{Percent of Correct Keypoints (PCK)} is an accuracy metric that determines whether the distance between the predicted keypoint and  the true joint location lies within a varying threshold.

 \section{Human Pose Estimation} \label{sec:hpe}

\noindent In this section, we review CNN-based human pose estimation techniques that operate on images (Section~\ref{hpi-from-images}) and videos (Section~\ref{hpi-from-videos}). The methods presented contribute to animal behavior classification frameworks presented in Section~\ref{sec:animalbehavior}. Table~\ref{tab:image-datasets-table} summarizes common benchmark image datasets used in deep learning, with pose estimation datasets indicated with bold text. The key contributions of the methods discussed are summarized in Table \ref{tab:pose-estimation-summary-table}. 

Human pose estimation methods generate a set of $(x, y)$ coordinates that correspond to the position of body joints within an image frame. These predictions are used as input to pose-based action recognition methods. Using taxonomy from deep learning-based human pose estimation~\cite{Zheng2020}, we categorize these approaches as \hbox{2D} or \hbox{3D}, and single-person or multi-person. The \hbox{2D} single-person methods are regression-based or detection-based. Regression-based methods regress an image to find joint locations, while detection-based methods detect body parts using heat maps. The \hbox{2D}  multi-person methods use top-down or bottom-up pipeline approaches. Top-down pipeline approaches use person detectors to obtain a set of bounding boxes for each person in the image before applying single-person pose estimation techniques to each bounding box to obtain the multi-person pose estimation. Bottom-up pipeline approaches first locate all the joints of all people in an image, and then assign them to individuals. We cover \hbox{3D} pose estimation from image or video input, and categorize methods as single-view/single-person, single-view/multi-person or multi-view.

\begin{table} [h]
\begin{center}
\begin{tabular}{|m{0.3\columnwidth} | m{0.075\columnwidth} | m{0.15\columnwidth} | m{0.1\columnwidth} | m{0.1\columnwidth} |}
\hline
Dataset & Year & Size & Classes & Joints \\
\hline
MNIST \cite{LeCun1998} & $1998$ & $70,000$ & $10$ & - \\ \hline
ImageNet \cite{Deng2009} & $2009$ & $1,431,167$ & $1,000$ & - \\ \hline 
CIFAR-10 \cite{Krizhevsky2009} & $2009$ & $60,000$ & $10$ & - \\ \hline
\textbf{HumanEva \cite{Sigal2010}} & $2009$ & $40,000$ & - & $15$ \\ \hline
\textbf{LSP \cite{Johnson2010}} & $2010$ & $2,000$ & - & $14$ \\ \hline
TFD~\cite{Susskind2010} & $2010$ & $112,234$ & $7$ & - \\ \hline
Cropped SVHN~\cite{Netzer2011} & $2011$ & $630,420$ & $10$ & - \\ \hline
\textbf{\hbox{LSP-extended~\cite{Johnson2011}}} & $2011$ & $10,000$ & - & $14$\\ 
\hline
\textbf{FLIC \cite{Sapp2013}} & $2013$ & $5,003$ & - & $10$ \\ 
\hline
\textbf{FLIC-motion \cite{JainTL2015}} & $2015$ & $5,003$ & - & $10$ \\ \hline
\textbf{MPII human pose} \cite{Andriluka2014} & $2014$ & $40,522$ & - & $16$\\ \hline
\textbf{Human3.6M} \cite{Ionescu2014} & $2014$ & $3.6M$ & - & $24$\\ \hline
\textbf{MPI-INF-3DHP \cite{Mehta2017}} & $2017$ & $1.3M$ & - & $24$ \\
\hline
\end{tabular}
\end{center}
\vspace{5pt}
\caption{Common Benchmark datasets for Image Classification and Pose Estimation from Images. The right columns indicate whether the data has class or joint labels.} \label{tab:image-datasets-table}
\end{table}

\begin{table} [h]
\begin{center}
\begin{tabular}{| m{1.6cm} | m{0.5cm} | m{1.2cm} | m{1cm} | m{1.1cm} | m{.8cm}|}
\hline
Dataset & Year & Trimmed & Size  & Average Length (sec.) & \hspace{-2pt} Classes \\
\hline
CCV \cite{Jiang2011} & $2011$ & \cmark & $9,317$ & $80$  & $20$ \\
\hline
HMDB51 \cite{Kuehne2011} & $2011$ & \cmark & $7,000$ & $\sim5$ & $51$ \\
\hline
UCF101 \cite{Soomro2012} & $2012$ & \cmark & $13.3K$ & $\sim6$ & $101$ \\
\hline
Sports-1M \cite{Karpathy2014} & $2014$ & \cmark & $1.1M$ & $\sim330$ & $487$ \\
\hline
THUMOS14 \cite{Jiang2014} & $2014$ & \xmark & $5,084$ & $233$ & $101$ \\
\hline
ActivityNet \cite{Heilbron2015} & $2015$ & \xmark & $9,682$ & $[300, 600]$  & $203$ \\
\hline
Kinetics-400 \cite{Kay2017} & $2017$ & \cmark & $306K$ & $10$  & $400$ \\
\hline
Kinetics-600 \cite{Carreira2018}& $2018$ & \cmark & $482K$ & $10$  & $600$ \\
\hline
\end{tabular}
\end{center}
\vspace{5pt}
\caption{Common Video Benchmark Datasets} 
\label{video-datasets-table}
\end{table}

\subsection{Pose Estimation from Images} \label{hpi-from-images}

\begin{table}
\begin{center}
\begin{tabular}{|c|} 
 \hline
 Input $64 \times 64$ color image \\ 
 \hline
 $5 \times 5$ conv. RELU. $2 \times 2$ MaxPool. $\mathbb{R}^{16 \times 32 \times 32}$ \\
 \hline
 $5 \times 5$ conv. RELU. $2 \times 2$ MaxPool. $\mathbb{R}^{32 \times 16 \times 16}$ \\
 \hline
 $5 \times 5$ conv. RELU. now $\mathbb{R}^{16 \times 32 \times 32}$. flatten to $\mathbb{R}^{8192}$\\
 \hline
 FC. RELU. $\mathbb{R}^{500}$\\ 
 \hline
 FC. RELU. $\mathbb{R}^{100}$\\ 
 \hline
 FC. logistic unit. $\mathbb{R}^{1}$\\
 \hline
\end{tabular}
\end{center}
\vspace{5pt}
\caption{Body Part Detector Architecture~\cite{JainTA2014} for Human Pose Estimation } 
\label{tab:jointdetector}
\end{table}

\vspace{10pt}
\textbf{\hbox{2D} Single-Person Pose Estimation}

\vspace{10pt}
\noindent\textbf{Detection-Based:} One of the first pose estimation techniques to apply a CNN was a two-stage filtering approach~\cite{JainTA2014} for finding joint positions. The first stage generates a binary response map for each joint, applying multiple CNNs as sliding windows on overlapping areas of an image.  The response map is a unary distribution representing the confidence of the joint's presence over all pixel positions. Denoising in the second stage removes false positives by passing the CNN output through a higher-level spatial model with body pose priors computed on the training set. Prior conditional distributions for two joints $(a, b)$ are calculated as a histogram of joint $a$ locations given that joint $b$ is at the image center. Similarly, a global position prior for the face is calculated using a histogram of face positions. A process similar to sum-product belief propagation~\cite{Pearl1982} generates the filtered distributions for each joint given unary distributions produced by the CNNs and the prior conditional distributions. 

This non-linear mapping from pixels to vector representations of articulated pose is challenging. Data loss during pooling, and high numbers of invalid poses in the training data make valid poses difficult to learn. To avoid this, a performance improvement trained multiple CNNs for independent binary joint detection, using one network for each feature. Each CNN (shown in Table \ref{tab:jointdetector}) has a single output unit representing the probability of a joint being present in the image patch. Results were evaluated on the FLIC~\cite{Sapp2013} dataset, which consists of images of Hollywood actors in front-facing poses. The percentage of correct joint predictions within a given precision radius (in pixels) is used as a quantitative metric for pose estimation accuracy. The model outperformed state-of-the-art estimation methods available at the time.

OpenPose uses a convolutional pose machine~\cite{Wei2016} to predict keypoint coordinates, and part affinity fields to determine correspondences between keypoints and subjects in the image. Convolutional Pose Machines are a sequence of CNNs that produce increasingly refined belief maps for estimating \hbox{2D} pose, without a graphical model. This design enforces intermediate supervision and limits vanishing gradients. Transfer learning with pre-training on ImageNet improves performance. Predefined rules are used to determine if pose features represent specific behaviors.

\vspace{10pt}
\noindent\textbf{Regression-based:} Human pose estimation may be formulated as a continuous regression problem. DeepPose~\cite{Toshev2014} regresses an image to a normalized pose vector representing body joint locations. The network input is ($\bm{X}$, $\bm{y}$), where $\bm{X}$ is a labeled image and $\bm{y} = \{..., y_i^T, ...\}^T$ is the true pose vector. Each $\bm{y}_i$ is the absolute image coordinates of the $i$th joint, where $i \in \{1,...,k\}$ and $k$ is the number of joints. Each joint position must be normalized with respect to a bounding box, $b$, centered around the human subject to produce a normalized pose vector $N(\bm{y}; b) = (..., N(\bm{y}_i; b)^T, ...)^T$.

The DeepPose CNN architecture $\psi$ is based on AlexNet~\cite{Krizhevsky2012}. However, rather than formulate a classification problem, a linear regressor is trained to learn the parameters $\bm{\theta}$ of a function $\psi(\bm{X}; \bm{\theta}) \in \mathbb{R}^{2k}$ such that an image $\bm{X}$ regresses to a pose prediction $\bm{y}^*$ in absolute image coordinates:
\begin{equation}
\bm{y}^* = N^{-1} (\psi (N(\bm{X}); \bm{\theta}))
\end{equation}

The linear regressor is trained on the last network layer to predict a pose vector by minimizing the $L_2$ distance between the prediction and ground truth pose vector. The training set is first normalized as explained, yielding the following optimization problem:
\begin{equation}
\underset{\bm{\theta}}{\mathrm{argmin}} \sum_{(\bm{X}, \bm{y}) \in D_N} \sum^k_{i = 1} \| \bm{y}_i - \psi_i ((\bm{X}, \bm{\theta}) \|^2_2,
\end{equation}

\noindent where $D_N$ is the normalized training set:
\begin{equation}
D_N = \{(N(\bm{X}), N(\bm{y})) \vert (\bm{X}, \bm{y}) \in D)\}.
\end{equation}

DeepPose CNNs operate on coarse scale images at a fixed resolution ($220 \times 220$ pixels), making it difficult to analyze fine image detail at the level required for precise joint localization. To address this, a cascade of pose regressors was introduced after the first stage to predict the displacement of the predicted joint locations (from prior states) to the true joint locations. Relevant parts of the image are targeted at higher resolutions for higher precision without increasing computation costs. The model was evaluated on datasets that include a variety of poses including the aforementioned FLIC dataset~\cite{Sapp2013}, the Leeds Sports Poses dataset (LSP)~\cite{Johnson2010} of athletes participating in sports, and its extension~\cite{Johnson2011}. Limb detection rates evaluated using PCP and PDJ produced state-of-the-art or better results on both metrics.

\vspace{10pt}
\noindent\textbf{Joint Training with A Graphical Model:} Hybrid methods that combine multiresolution CNNs and joint training with graphical models produce higher accuracy rates than DeepPose. These models address limitations with network capacity, overfitting, and inefficiencies caused by nonlinear mappings from image to vector space.  CNNs~\cite{Zhang2020} that operate on multi-resolution input, can identify a broad range of feature sizes in a single forward pass by adapting the effective receptive field within CNN layers. 

Tompson et. al.~\cite{TompsonJ2014} combined a CNN with a Markov Random Field, an undirected graph with Markov properties~\cite{Murphy2012}.  A multi-resolution feature representation with overlapping receptive fields was used to perform heat-map likelihood regression. Starting with an input image, a sliding window CNN generates a per-pixel heat-map of likelihoods for joint locations. The CNN was trained jointly with a graphical model. In later stages, a spatial model predicts which heat-maps contain false positives and incorrect poses. This is done by constraining joint inter-connectivity and enforcing global pose consistency. The graph is learned implicitly without hand-crafted pose priors or graph structure~\cite{JainTA2014}.  The results showed improved accuracy compared to the previous state-of-the-art~\cite{JainTA2014,Toshev2014} in human body pose recognition using the FLIC and LSP-extended datasets with PDJ as the evaluation metric. 

A later extension~\cite{TompsonG2015} improved localization accuracy by recovering the precision lost due to pooling. The input image was first passed through a CNN for coarse-level pose estimation to produce initial heat maps with per-pixel likelihoods of joint positions. The heat-maps were then passed through a position refinement model to refine the pose prediction. This modification outperformed the previous state-of-the-art on the FLIC and MPII human pose~\cite{Andriluka2014} benchmarks. The MPII human pose dataset is comprised of $24,520$ images of humans performing everyday activities collected from YouTube videos. This method outperformed previous state-of-the-art pose estimation methods~\cite{JainTA2014,Toshev2014,TompsonJ2014} on the FLIC dataset using PCK scores. Another method uses a \emph{Graphical Model with Image Dependent Pairwise Relations}~\cite{Chen2014} to generate input for both the unary and pairwise terms from image patches. Experiments performed on the FLIC and LSP datasets showed that this formulation improves upon DeepPose according to multiple metrics.

\begin{table*} [b]
\begin{center}
\begin{tabular}{| m{1.25cm} | m{0.5cm} | m{0.75cm} | m{13cm} |} 
 \hline
 Method & Year & 2D/3D & Key Contribution(s)  \\ 
 \hline
\cite{JainTA2014} & 2014 & 2D &
        \begin{itemize} 
            \item The development of a deep CNN architecture to learn low-level features and a higher-level spatial model.
        \end{itemize} \\ 
 \hline
\cite{TompsonJ2014} & 2014 & 2D &
        \begin{itemize} 
            \item The development of a hybrid architecture that incorporates a deep CNN and a Markov Random Field that can exploit geometric constraints between body joint locations.
        \end{itemize} \\ 
 \hline
\cite{Chen2014} & 2014 & 2D &
        \begin{itemize} 
            \item The development of a graphical model that uses pairwise relations to exploit how local image measurements can be used to predict joint locations and relationships.
            \item The use of a deep CNN to learn conditional probabilities for the presence of joints and their spatial relationships in image patches.
        \end{itemize} \\ 
 \hline
\cite{Li2015} & 2015 & 3D &
        \begin{itemize} 
            \item The application of a deep CNN to 3D pose estimation from monocular images.
            \item The joint training of pose regression and body part detectors.
        \end{itemize} \\ 
 \hline
\cite{Pfister2015pose} & 2015 & 2D &
        \begin{itemize} 
            \item The use of a deep CNN for estimating human pose in videos.
            \item The use of temporal information between frames to improve performance.
        \end{itemize} \\ 
 \hline
DeepPose \cite{Toshev2014} & 2014 & 2D &
        \begin{itemize} 
            \item The formulation of human pose estimation as a deep CNN-based regression problem.
            \item The use of a cascade of regressors to increase precision.
        \end{itemize}\\ 
\hline
MoDeep \cite{JainTL2015} & 2015 & 2D &
        \begin{itemize} 
            \item The use of motion features for human pose estimation.
        \end{itemize} \\ 
\hline
\cite{TompsonG2015} & 2015 & 2D &
        \begin{itemize} 
            \item The introduction of a position refinement model that is trained to estimate the joint offset location in a local image region in order to improve joint localization accuracy.
        \end{itemize} \\ 
\hline
DeepCut \cite{Pishchulin2016deepcut} & 2016 & 2D &
        \begin{itemize} 
            \item The development of a partitioning and labeling formulation of body-part predictions produced by CNN-based joint detectors.
            \item This formulation infers the number people in images, identifies occlusions, and differentiates between overlapping body parts of nearby people. 
        \end{itemize} \\ 
\hline
\cite{Li2017}  & 2017 & 3D &
        \begin{itemize} 
            \item The modeling of dependencies between joints using a max-margin structured learning framework for monocular 3D pose estimation.
        \end{itemize} \\ 
 \hline
\cite{Tekin2016a} & 2016 & 3D &
        \begin{itemize} 
            \item The use of an overcomplete  autoencoder for monocular 3D pose estimation that learns a latent pose representation and models joint dependencies.
        \end{itemize} \\ 
 \hline
\cite{Tekin2016b} & 2016 & 3D &
        \begin{itemize}
            \item The use of motion information from video clips for 3D pose estimation and the direct regression from clips to pose in the center frame.
        \end{itemize} \\ 
 \hline
\cite{Pavlakos2017} & 2017 & 3D &
        \begin{itemize} 
            \item The discretization of 3D space around a subject for monocular 3D pose estimation.
            \item The use of a CNN to predict per voxel likelihoods for each joint.
        \end{itemize} \\ 
 \hline
DeeperCut \cite{Insafutdinov2016} & 2016 & 2D &
        \begin{itemize} 
            \item The design of very deep body part detectors by building upon the ResNet architecture.
            \item The use of image conditioned pairwise terms between body parts to improve performance for images with multiple people.
            \item The development of an incremental optimization procedure that leads to speed and performance boosts.
        \end{itemize} \\ 
\hline
\end{tabular}
\end{center}
\vspace{5pt}
\caption{Summary of Pose Estimation Approaches} 
\label{tab:pose-estimation-summary-table}
\end{table*}

\begin{table*} [h]
\begin{center}
\begin{tabular}{| m{0.5cm} | m{2cm} | m{0.5cm} | m{1.8cm} | m{10cm} |} 
 \hline
 CNN Arch. & Method & Year & Classification Accuracy & Key Contribution(s)  \\ 
 \hline
\rotatebox[origin=c]{90}{2D} &\hbox{Slow Fusion} \cite{Karpathy2014} & $2014$ & $65.4\%$ & 
        \begin{itemize} 
            \item The release of the Sports-1M dataset.
            \item The evaluation of CNNs for video classification and the development of slow fusion, late fusion, and early fusion.
        \end{itemize} \\ 
 \hline
\rotatebox[origin=c]{90}{Two-Stream} & Two-Stream CNN \cite{Simonyan2014} & $2014$ & $88.0\%$ & 
        \begin{itemize} 
            \item The development of the two-stream architecture that incorporates spatial and temporal networks. 
            \item The usage of optical flow to model movement in videos for action recognition. 
        \end{itemize} \\ 
 \cline{2-5}
 & MotionNet \cite{Zhu2019} & $2019$ & $89.82\%$ & 
        \begin{itemize} 
            \item The development of the MotionNet CNN that implicitly calculates optical flow between frames as a part of a two-stream architecture.
        \end{itemize} \\ 
 \hline
 \rotatebox[origin=c]{90}{3D}& C3D \cite{Tran2015} & $2015$ & $86.7\%$ & 
        \begin{itemize} 
            \item The development of an $8$-layer 3D CNN that operates over $16$ frames and uses filters of size $3 \times 3 \times 3$.
        \end{itemize} \\ 
 \cline{2-5}
 & $R(2+1)D$ \cite{Tran2018} & $2018$ & $97.3\%$ & 
        \begin{itemize} 
            \item The factorization of 3D convolution into 2D spatial convolution followed by 1D temporal convolution.
        \end{itemize} \\ 
 \hline
\rotatebox[origin=c]{90}{Two-Stream/3D} & Two-Stream \hbox{CNN Fusion} \cite{Feichtenhofer2016} & $2016$ & $93.5\%$ & 
        \begin{itemize} 
            \item The development of a two-stream architecture that uses 3D convolutional fusion and 3D pooling.
        \end{itemize} \\
 \cline{2-5}
& I3D \cite{Carreira2017} & $2017$ & $98.0\%$ & 
        \begin{itemize} 
            \item The usage of the Kinetics-400 dataset \cite{Carreira2017}.
            \item The development of two-stream inflated 3D CNNs that expand the filters and pooling kernels of very deep image CNNs into 3D. 
        \end{itemize} \\ 
 \hline 
\end{tabular}
\end{center}
\vspace{5pt}
\caption{Summary of Discriminative Action Recognition Approaches and their Performance on UCF101~\cite{Soomro2012}} 
\label{tab:action-recognition-summary-table}
\end{table*}

\vspace{10pt}
\textbf{\hbox{2D} Multi-Person}

\vspace{10pt}
There are additional challenges with pose estimation when multiple subjects interact in a video. Most early strategies for multi-person pose estimation used a two-stage inference process to first detect and then independently estimate poses. However, this method is less effective when multiple people have overlapping body parts because the same body part candidates are often assigned to multiple people.

A proposed solution, DeepCut~\cite{Pishchulin2016deepcut}, casts the joint detection and pose estimation problems as integer linear programs. DeepCut's formulation is a Joint Subset Partitioning and Labeling Problem (SPLP) that jointly infers the number of people and poses, the spatial proximity, and areas of occlusion. The solution jointly estimates the poses of all people in an image by minimizing a joint objective. A set of joint candidates is partitioned and labeled into subsets that correspond to mutually consistent joint candidates that satisfy certain constraints. Even though the problem is NP-hard, this formulation allows feasible solutions to be computed within a certain optimality gap. DeepCut adapts two CNN architectures, Fast R-CNN~\cite{Girshick2015} and VGGNet~\cite{Simonyan2015}, to generate body part candidates. Significant improvement was shown over previous methods~\cite{TompsonJ2014,Chen2014,TompsonG2015} for both single and multi-person pose estimation according to multiple evaluation metrics on the LSP, LSP-extended, and MPII human pose datasets.

DeeperCut~\cite{Insafutdinov2016} doubled the pose estimation accuracy of DeepCut while reducing the running time by $2$-$3$ orders of magnitude. Contributions included novel image conditioned pairwise terms between body parts that improved performance for images with multiple people, and a novel optimization method that decreased runtime while improving pose estimation accuracy. DeeperCut adapted ResNet~\cite{He2016}, which is $8 \times$ deeper than VGGNet. Experiments were conducted on both single and multiple person human pose datasets: LSP, LSP-extended, and MPII using the PCK evaluation metric. Results yielded a new state-of-the-art in multi-person~\hbox{2D} pose estimation.

\vspace{10pt}
\textbf{\hbox{3D} Pose Estimation}

\vspace{10pt}
Estimating a~\hbox{3D} human pose from a single RGB image is a challenging problem in computer vision because it requires solving two ambiguous tasks~\cite{Tome2017}. 
The first task is finding the \hbox{2D} location of human joints in the image. This is hard due to different camera viewpoints, occlusions, complex body shapes, and varying illumination. The second, converting the coordinates of \hbox{2D} landmarks into \hbox{3D} coordinates, is an ill-posed problem that requires additional information such as \hbox{3D} geometric priors and other constraints.  Methods for \hbox{3D} pose inference from images either regress the \hbox{3D} pose directly from images, or first estimate the~\hbox{2D} pose and then lift the coordinates into \hbox{3D} using a pipeline approach. Three-dimensional human pose estimation approaches presented in the remainder of this subsection are single-view/single-person.

\vspace{10pt}
\noindent\textbf{Regression-based:} CNNs can directly regress~\hbox{3D} poses from images. One method, \emph{\hbox{3D} Human Pose Estimation from Monocular Images}~\cite{Li2015}, jointly trains the pose regression task with a set of recognition tasks in a heterogeneous multi-task learning framework.  The network was pre-trained using the recognition tasks, then refined using only the pose regression task. A \emph{Maximum-Margin Structured Learning} technique~\cite{Li2017} takes an image and a \hbox{3D} pose as input, and produces a score that indicates whether the pose is depicted. During training a maximum-margin cost function is used to enforce a re-scaling margin between the score values of the ground truth image-pose pair and the rest of the pairs. The results showed improved performance. Overcomplete autoencoders~\cite{Tekin2016a} have been used to learn a high-dimensional latent pose representation. This method accounts for joint dependencies that predict \hbox{3D} human poses from monocular images. The problem was also posed as a key point localization problem in a discretized \hbox{3D} space~\cite{Pavlakos2017}. In this case, a CNN is trained to predict per voxel likelihoods for each joint in the volume.

At the time, among pipeline approaches, it was unclear whether the remaining error in state-of-the-art methods was due to a limited~\hbox{2D} pose understanding, or from a failure to map~\hbox{2D}  poses to~\hbox{3D} positions. To better understand the sources of error, the~\hbox{3D} human pose estimation problem was decoupled into the well-studied problems of~\hbox{2D} pose estimation from images and~\hbox{3D} pose estimation from~\hbox{2D} joint detections. 

\vspace{10pt}
\noindent\textbf{\hbox{2D} to \hbox{3D} Lifting:} By focusing on the latter problem, ground truth~\hbox{2D} joint locations can be~\emph{lifted} to ~\hbox{3D}  space using a relatively simple deep CNN.  This yielded state-of-the-art results on the Human3.6M dataset~\cite{Ionescu2014} which consists of $3.6$ million~\hbox{3D}  human poses. The findings indicated that the remaining error in pipeline approaches for~\hbox{3D} pose estimation techniques stem from~\hbox{2D} pose analysis. Another pipeline approach~\cite{Tome2017} used a pre-learned~\hbox{3D} human pose model as part of the CNN architecture itself. The architecture learned to use physically plausible~\hbox{3D} reconstructions in its search for better~\hbox{2D} landmark locations. The process obtained state-of-the-art~\hbox{2D} and~\hbox{3D} results on Human3.6M, and demonstrated the importance of considering~\hbox{3D}, even when solving~\hbox{2D}  pose estimation problems.

\vspace{10pt}
\noindent \textbf{Generalization to in-the-wild conditions:}
The previous pose estimation methods had low generalization to in-the-wild conditions. This is due to inherent limitations in the training datasets. A \emph{Monocular \hbox{3D} Human Pose Estimation} method~\cite{Mehta2017} used transfer learning to leverage the mid and high-level features learned on existing~\hbox{2D}  and~\hbox{3D} pose datasets, yielding improved performance. State-of-the-art results were obtained on the Human3.6M dataset and the HumanEva pose estimation benchmark,~\cite{Sigal2010}. A new benchmark dataset, MPI-INF-3DHP \cite{Mehta2017}, of indoor and outdoor scenes was presented. Transfer learning with MPI-INF-3DHP and~\hbox{2D} pose data yielded the best generalization to in-the-wild conditions, confirming that transfer learning is beneficial for~\hbox{3D} pose estimation.

\subsection{Human Pose Estimation from Video}\label{hpi-from-videos}

\begin{figure}[ht]
\centering
\includegraphics[width=0.4\textwidth]{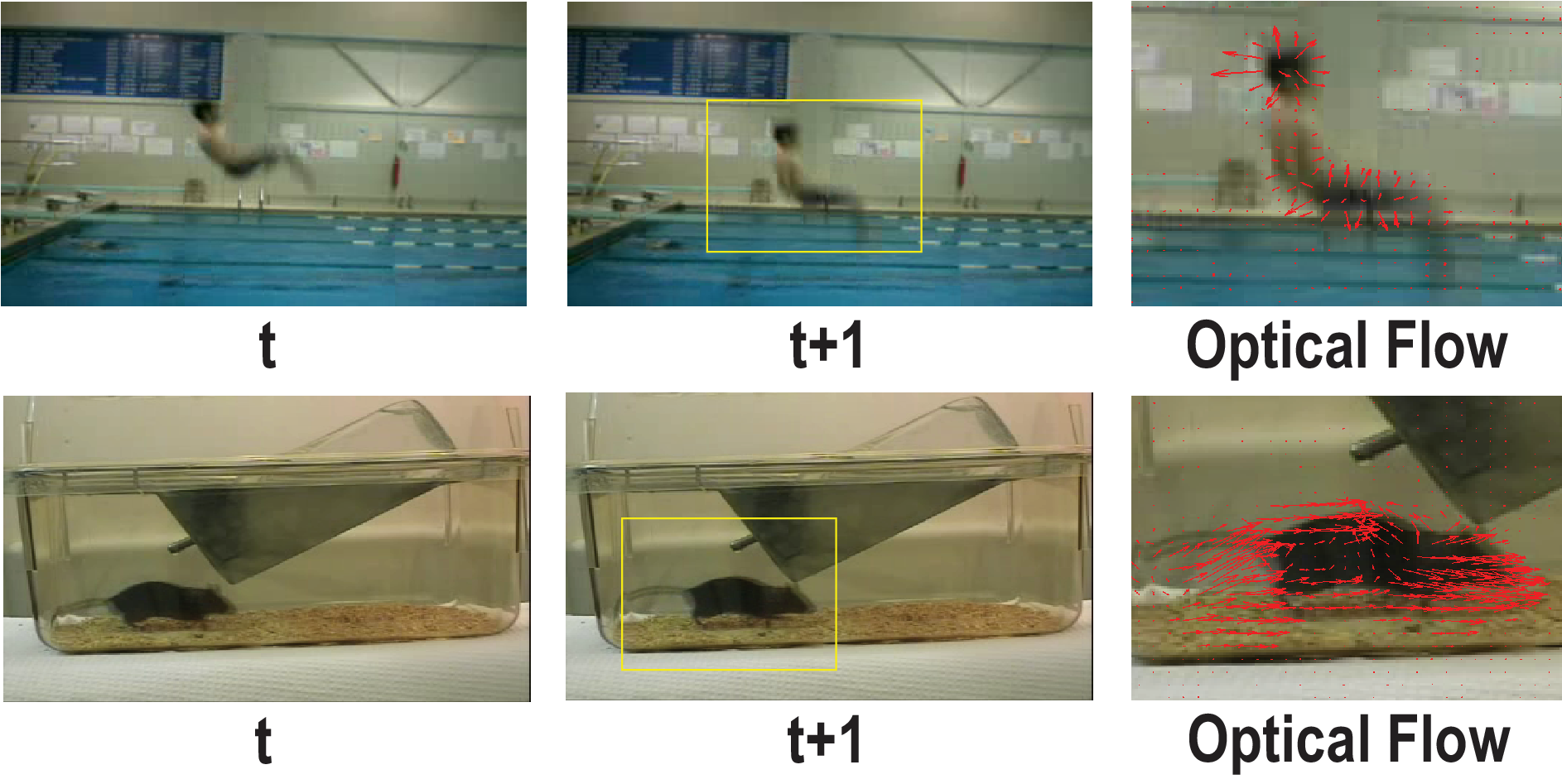}
    \caption{(top row) Two consecutive frames and a close-up from a video of a person diving from the UCF101~\cite{Soomro2012}. (bottom row) Two consecutive frames and a close-up from a video of a mouse with a close-up from a published~\cite{Jhuang2010} animal behavior dataset.  In each case, optical flow between the frames was calculated using MATLAB and superimposed on the second frame and close-up.}
    \label{flow-figure}
\label{fig:trimmed-figure2}
\end{figure}

Human pose estimation approaches that operate on videos handle temporal information using techniques similar to those we cover in our discussion of human action recognition methods (Section~\ref{sec:actionrecognition}). CNNs used for action recognition initially incorporated temporal information using optical flow~\cite{Brox2004}  as an input motion feature~\cite{Simonyan2014} to a two-stream network architecture (Sections~\ref{sec:motivation} and~\ref{sec:actionrecognition}).  An optical flow~\cite{Brox2004} (shown in Figure~\ref{fig:trimmed-figure2}), is a set of displacement vector fields $\bf{d}_t$ between pairs of consecutive frames $t$ and $t + 1$. Figure~\ref{flow-figure} (top row) shows an example of the optical flow between two consecutive frames in a video of a person diving. Figure~\ref{flow-figure} (bottom row) shows a mouse walking. 

Two \hbox{2D} single-view/single-person pose estimation methods~\cite{Pfister2015pose,JainTA2014}, one that is detection-based and another that is regression-based, extended this idea using optical-flow maps from multiple nearby frames to predict the pose in the current frame. In a related  work, \emph{Flowing ConvNets for Human Pose Estimation}~\cite{Pfister2015flow}, when performing inference for one frame, joint positions for all neighboring frames are explicitly predicted and aligned to the corresponding frames. This is done using a dense optical flow to warp backward or forward in time. Evaluation performed on three large pose estimation datasets showed that this method outperforms the previous state-of-the-art.  

MoDeep~\cite{JainTL2015}, and similar variants~\cite{Pfister2015pose}, improved upon DeepPose by exploiting motion as a cue for body part localization. This single-view/single-person method uses optical flow and color from multiple nearby frames as input to a multi-resolution CNN. The multi-resolution CNN was designed for estimating a human's pose in a video given an image and a set of motion features. Accuracy in high-precision regions was improved (compared to DeepPose) by incorporating motion features, and using a translation-invariant model. Optical flow was one of several formulations considered. The FLIC-motion \cite{JainTL2015} dataset which consists of the original FLIC dataset \cite{Sapp2013} augmented with motion features, was created. MoDeep performed best on FLIC-motion compared to other state-of-the-art pose estimation methods. A related technique by Tekin et. al. ~\cite{Tekin2016a} extended a CNN architecture for pose estimation on still images~\cite{Tekin2016b} to operate on video. It included temporal information in the joint predictions by extracting spatio-temporal information from a sequence of frames.

\section{Human Action Recognition}\label{sec:actionrecognition} 
\begin{figure}[h]
\begin{center}
\includegraphics[width=\linewidth]{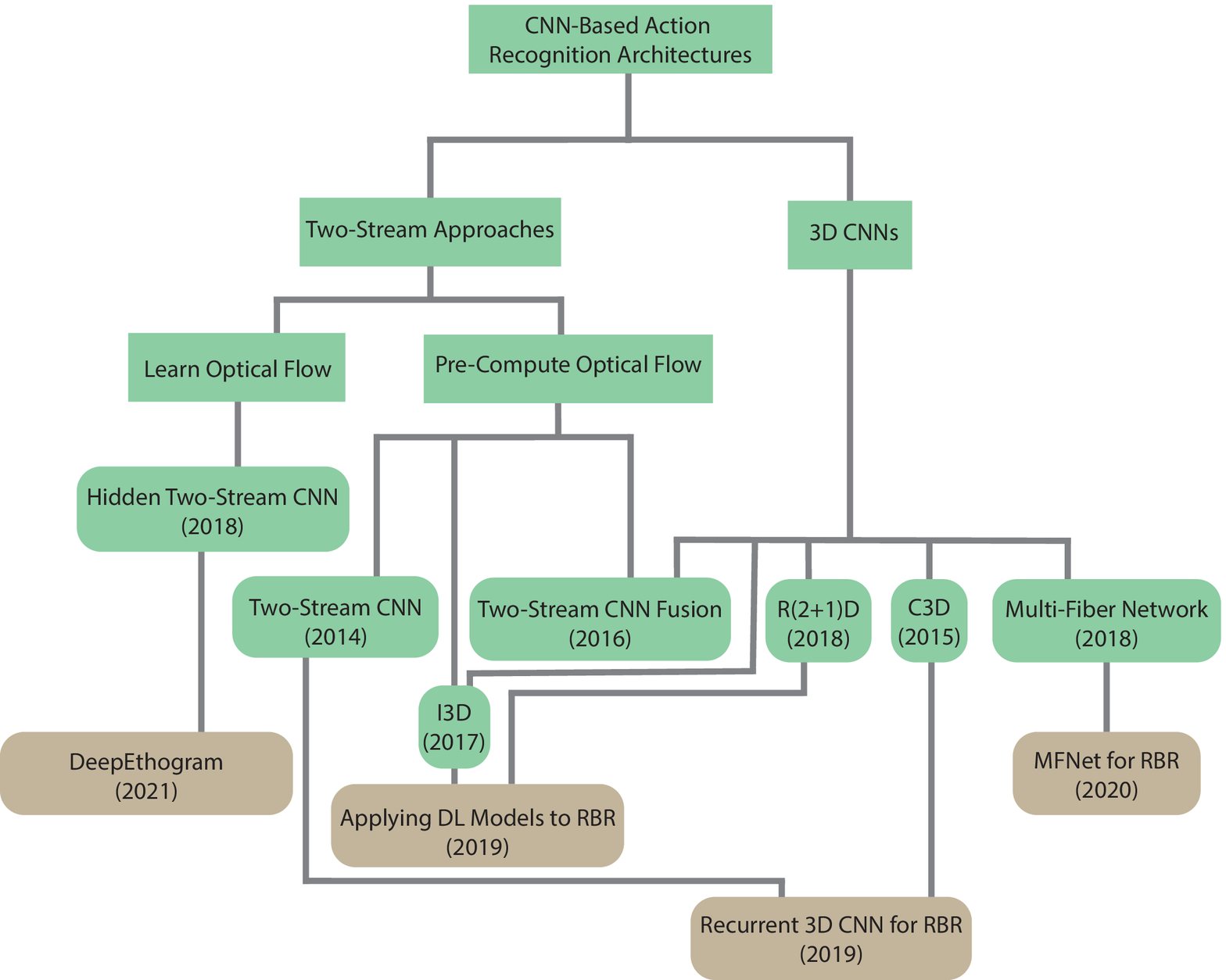}
\caption{CNN-based Action Recognition Architectures for Human Actions (green) and Animal Behavior Classification (brown). Rectangles denote architecture categories and rounded rectangles denote architectures.}
\label{fig:actionrecognition}
\end{center}
\end{figure}

Now that we have introduced CNN-based pose estimation techniques, we turn our attention to action recognition methods that track motion across video frames. Figure~\ref{fig:actionrecognition} illustrates the CNN architectures discussed in this section in green. Rectangles denote architecture categories and rounded rectangles denote architectures. Architecture we discuss have been extended for animal behavior classification systems (shown in brown), that we review in Section~\ref{sec:animalbehavior}.

\textbf{Standard Action Recognition Methods} extract features, sparse \cite{Laptev2003} or dense~\cite{Wang2011}  spatio-temporal interest points, for example, and then track their displacement through frame sequences using techniques like optical flow. Improved Dense Trajectories (iDT)~\cite{Wang2013} is an example that represents motion as feature trajectories, fixed-sized descriptors computed along the paths of feature points. In this case, quantization using a k-means dictionary~\cite{Laptev2008} is used to combine features. A sparse vector with stored occurrence counts of a vocabulary of local image features (bag of visual words), is then accumulated across the video. A support vector machine (SVM), or other classifier, is trained to classify the bag of visual words to discriminate video-level classes.

\textbf{CNN-Based Temporal Learning:} Modern deep learning methods use neural networks to detect spatio-temporal features. In our survey, we focus on CNN-based examples. While images can be scaled and cropped, and are easily processed with fixed-sized CNNs, videos are difficult to work with because they vary in temporal content. According to Karpathy et. al.~\cite{Karpathy2014}, compared to traditional machine learning approaches for action recognition~\cite{Laptev2003,Wang2011,Wang2013}, CNN-based solutions shift the engineering from feature design and extraction, to network architecture design and hyperparameter tuning.

\begin{figure}[h!]
\begin{center}
\includegraphics[width=\linewidth]{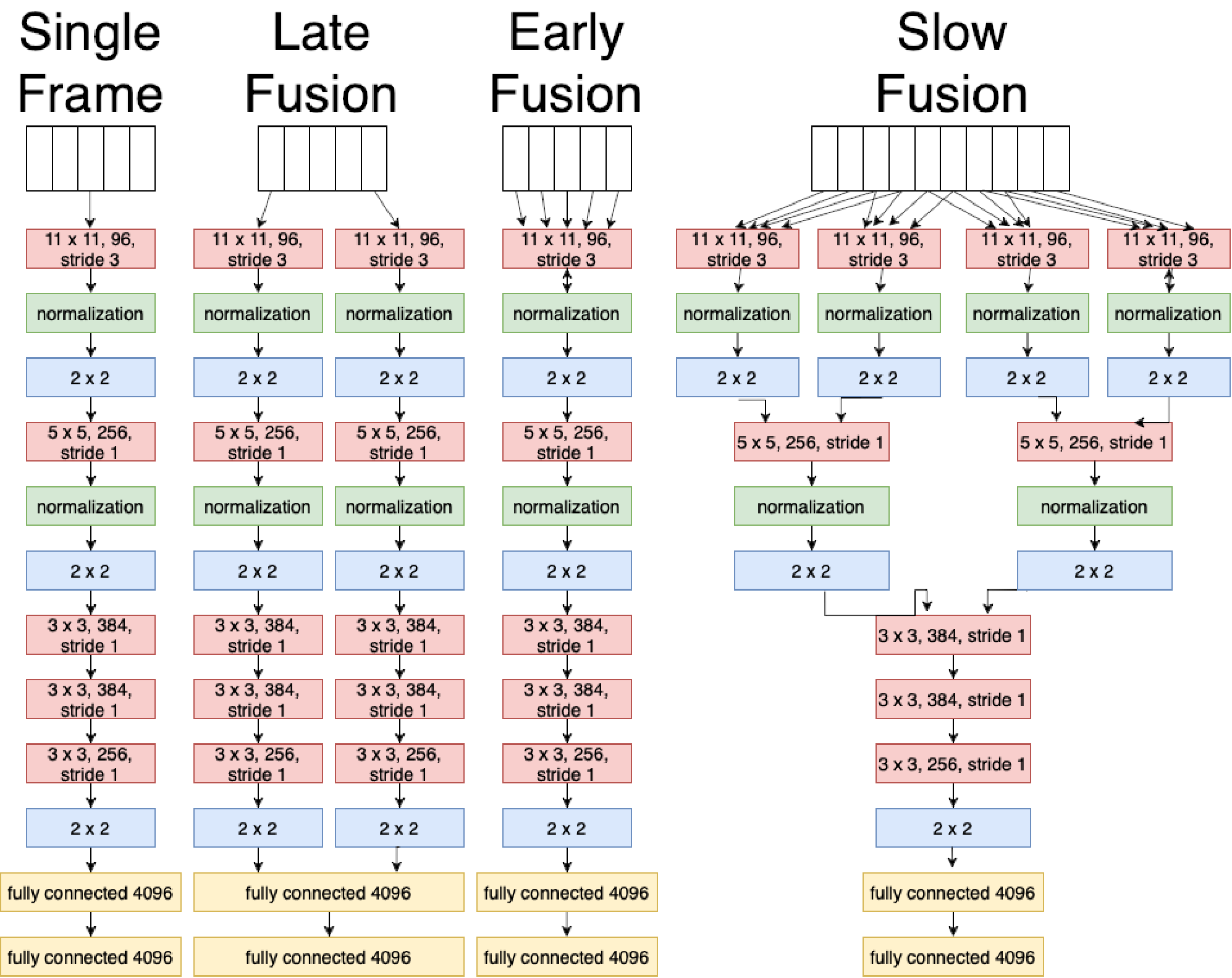}
\caption{Approaches~\cite{Karpathy2014} tested for fusing information in the temporal domain. Red, green, blue, and yellow layers denote convolution, normalization, pooling, and fully connected layers, respectively. Zoom in to see details. Adapted from~\cite{Karpathy2014}.} \label{fig:Karpathy-fig}
\end{center}
\end{figure}

We begin with a seminal paper by Karpathy et. al.~\cite{Karpathy2014} that extended the connectivity of a CNN to operate in the temporal domain for video classification. The solution treats the input data as a bag of fixed-sized clips where one hundred half-second clips were randomly sampled from each video. Fusion operations are then used to extend the CNNs over the time dimension so that they operate on multiple frames and have the ability to learn spatiotemporal features from the clips. The team proposed three novel CNN connectivity patterns to fuse temporal information: Late fusion places two single-frame networks some time apart, then fuses their outputs later. Early fusion modifies the first layer to use a convolutional filter that extends in time. Slow fusion is a balance between early and late fusion (Figure \ref{fig:Karpathy-fig}). A baseline single-frame architecture was also used to analyze the contribution of static appearance to classification accuracy.

High volume video datasets with high class variation are essential for training and testing these new architectures. The largest video datasets available at the time were CCV~\cite{Jiang2011} which contains $9,317$ videos and $20$ classes,  and UCF101~\cite{Soomro2012} which contains $13,320$ videos and $101$ classes. Though effective for generating early action recognition benchmarks, the scale and variety of these video datasets were not equivalent to large image datasets like ImageNet \cite{Deng2009} (Section~\ref{sec:background}). To address this issue, the paper introduced a new benchmark action recognition dataset, the Sports-1M dataset~\cite{Karpathy2014} with one  million trimmed YouTube videos belonging to $487$ classes of sports. Each video is five minutes and $36$ seconds long on average. One hundred half-second clips were randomly sampled from each video in the Sports-1M dataset. To produce video-level predictions, $20$ clips from a video were randomly sampled and presented to each CNN individually and class predictions for each clip were averaged.

Slow fusion was the best fusion method for video classification on Sports-1M. Interestingly, the single-frame connectivity pattern performed similarly to the other patterns, which showed that learning from a series of frames showed no improvement compared to learning from a static image. The CNN's learned feature representation trained on Sports-1M was shown to generalize to different action classes and datasets. One transfer learning experiment using UCF101 compared the accuracy of training the CNN from scratch with the accuracies of fine-tuning the top layer, top three layers, and all layers. Fine-tuning the top three layers performed the best, demonstrating that the layers near the top learn dataset-specific features and the lower layers learn more generic features.

\textbf{Two-stream CNNs}~\cite{Simonyan2014} capture motion by decomposing videos into a spatial component, which has information about objects and scenes, and a temporal component, which describes the movement of objects and the camera. This approach is based on the two-streams hypothesis from neuroscience~\cite{Goodale1992} which posits that the human visual cortex has a ventral stream that performs object recognition, and a dorsal stream that recognizes motion.

The authors of a new two-stream CNN approach \cite{Simonyan2014} for video classification suggested that features learned through slow fusion  in the prior example~\cite{Karpathy2014} are not optimal for capturing motion because there is no significant difference between CNN performance on static images, and CNN performance using fusion. In addition, slow fusion was $20\%$ less accurate than the best trajectory-based shallow representation at the time~\cite{Peng2016}. The team proposed a solution that combined spatial and temporal recognition streams using late fusion. The temporal stream input is formed by stacking the optical flow displacement fields between a sequence of consecutive frames. Two input configurations were considered. One method does not sample optical flow at the same location across frames, but instead samples the optical flow along the motion trajectories. The other, more successful option, treats the horizontal and vertical components of the optical flow vector field $\bf{d}^x_t$ and $\bf{d}^y_t$, respectively, at frame $t$, as image channels. The optical flow channels of $L$ consecutive frames are stacked to form $2L$ input channels.
The CNN input volume $I_{\tau} \in \mathbb{R}^{w \times h \times 2L}$ for a frame $\tau$ is defined as:

\begin{equation}
I_\tau(u, v, 2k - 1) = d^x_{\tau + k - 1}(u, v), 
\label{equ:twostream-inputconfig1}
\end{equation}
\begin{equation*}
I_\tau(u, v, 2k) = d^y_{\tau + k - 1}(u, v),
\end{equation*}

\noindent where $w$ and $h$ are the width and height of the video, and the displacement vector at point $(u, v)$ in frame $t$ is $\bf{d_t}$$(u, v)$ with $u = [1;w]$ and $v = [1;h]$.  The frame and channel indices are $k = [1;L]$ and $c = [1;2L]$, respectively. The channels $I_\tau(u, v, c)$ represent the horizontal and vertical components of motion of a point $(u, v)$ over $L$ frames. From each displacement field $\bm{d}$, the mean vector is subtracted to perform zero-centering of the network input so that the optical flow between frames is not dominated by a particular displacement caused by camera motion.

This method was evaluated using two datasets: UCF101 and the Human Motion Database (HMDB51)~\cite{Kuehne2011}. HMDB51 has $6,766$ labeled video clips with $51$ classes. The spatial stream CNN was pre-trained on ImageNet. The best performance was recorded when stacking $L = 10$ horizontal and vertical flow fields and using mean subtraction. This method is not end-to-end trainable because a preprocessing step is required to compute optical flow~\cite{Brox2004} before training. The results established a new state-of-the-art for action recognition. 

Even with these improvements, two challenges remained. Two stream approaches developed thus far did not leverage correspondences between spatial cues and temporal cues (optical flow) in videos because fusion is only applied to classification scores. They also had a restricted temporal scale because the temporal CNN operates on a stack of $L$ consecutive optical flow frames. An improved two-stream architecture by Feichtenhofer et. al.~\cite{Feichtenhofer2016} addresses these issues by fusing spatial and temporal information more effectively across different levels of feature abstraction granularity. For example, the system fuses components like motion (recognized by the temporal network),  and location (recognized by the spatial network) to discriminate between actions like brushing teeth and brushing hair. The network is not able to distinguish between these actions when motion or location information is considered separately. The architecture was able to fuse  the spatial and temporal networks so that  temporal and spatial features correspond (spatial registration). The second limitation is addressed by capturing short-scale optical flow features using the temporal network and putting the features into context over a longer temporal scale at a higher layer. 

Fusion functions can be applied at different points in a traditional or two-stream CNN network to implement the fusion approaches we have described. For two feature maps $\bm{x}^a_t \in \mathbb{R}^{H \times W \times D}$ and $\bm{x}^b_t \in \mathbb{R}^{H' \times W' \times D'}$ at time $t$, a fusion function $f : \bm{x}^a_t, \bm{x}^b_t \rightarrow \bm{y}_t$ produces an output feature map $\bm{y}^t \in \mathbb{R}^{H'' \times W'' \times D''}$, where $H$, $W$ and $D$ are the height, width and number of channels. The one constraint is that $\bm{x}^a_t$ and $\bm{x}^b_t$ have the same spatial dimensions, so we assume $H = H' = H''$, $W = W' = W''$, $D = D'$.

The following functions were proposed to register corresponding pixels in the spatial and temporal stream networks. \textbf{\emph{Sum fusion}} computes the sum of the feature maps at the same spatial locations and channels. \textbf{\emph{Max fusion}} computes the maximum of the feature maps. \emph{\emph{Concatenation fusion}} stacks the feature maps at the same spatial locations across different channels using Equation~\ref{equ:concatenation-fusion}, where $\bm{y} = \mathbb{R}^{H \times W \times 2D}$. \textbf{\emph{Convolutional fusion}} first stacks the feature maps  using concatenation fusion, then convolves the stacked data with a bank of filters $\bm{f} \in \mathbb{R}^{1 \times 1 \times 2D \times D}$ accounting for biases $b \in \mathbb{R}^{D}$.  When used as a trainable filter kernel, $\bm{f}$ learns correspondences of feature maps that minimize a joint loss function. We include the formulation for convolutional fusion in Equation~\ref{equ:convolution-fusion} (with the intermediate concatenation step Equation~\ref{equ:concatenation-fusion}) as it has been found to be the most effective of these fusion methods.

\begin{equation} 
y^{cat}_{i, j, 2d} = x^{a}_{i, j, d} \ \ y^{cat}_{i, j, 2d - 1} = x^{b}_{i, j, d}
\label{equ:concatenation-fusion}
\end{equation}

\begin{equation}
\bm{y}^{conv} = \bm{y}^{cat} * \bm{f} + b.
\label{equ:convolution-fusion}
\end{equation}

\noindent \textbf{\emph{Bilinear fusion}} computes a matrix outer product of the feature maps at each pixel, then performs a summation over the pixels. The result captures multiplicative interactions between pixels in a high dimensional space. For this reason, in practice, bilinear fusion is applied after the fifth ReLU activation function and fully connected layers are removed. Ultimately each channel of the spatial network is combined multiplicatively with each channel of the temporal network.

As previously mentioned, convolution fusion was the best approach. When evaluating this fusion method, it was determined that additional computation speed-ups could be achieved by first combining feature maps over time using pooling methods before convolving with the filter bank. This effectively reduced the parameter space by reducing the feature map size. Two pooling scenarios were considered. Network predictions were averaged over time (\hbox{2D} pooling), or max pooling was applied to stacked data in a~\hbox{3D} cube (~\hbox{3D} pooling). Three-dimensional pooling increased performance more than~\hbox{2D} pooling. The network was fused spatially and truncated after the fourth convolutional layer to find a balance between maximizing classification accuracy and minimizing the total number of parameters. The fused architecture showed improvement upon previous state-of-the-art action recognition methods~\cite{Simonyan2014,Tran2015} on UCF101 and HMDB51.

Another challenge with two-stream approaches developed thus far, is the increased time and storage requirements due to their reliance on pre-computed optical flow before CNN training~\cite{Simonyan2014}. MotionNet~\cite{Zhu2019} introduced new time and storage efficiencies with a CNN that produced optical flow from video clips on-the-fly. The results were similar to the optical flow generated by one of the best traditional methods~\cite{ Zach2007}. Optical flow estimation was framed as an image reconstruction problem. Using two consecutive frames $\bm{I}_1$ and $\bm{I}_2$ as input, MotionNet generated a flow field $\bm{V}$ which was used with $\bm{I}_2$ to create a reconstructed frame $\bm{I}_1'$ using the inverse warping function~\cite{Jaderberg2015}.

MotionNet aims to minimize the photometric error between $\bm{I}_1$ and $\bm{I}_1'$. Assuming $\bm{I}_1$ and $\bm{I}_1'$ are related by a parametric transformation, the minimization function finds the parameters of the transformation $\bm{p}$ that minimize the squared intensity error:

\begin{equation} \label{photo-loss}
\underset{\bm{p}}{\mathrm{argmin}} \sum_{\bm{u} \in \alpha_1 }\frac{1}{2} \| \bm{I}_1(\bm{u}) - \bm{I}_1'(w(\bm{u}; \bm{p})) \|^2
\end{equation}
where $\bm{u} \in \alpha_1$ is a subset of pixel coordinates in image $\bm{I}_1$ and $w$ is the warping function~\cite{Baker2004}.
Traditionally the Lucas-Kanade algorithm~\cite{Lucas1981,Baker2004} solves Equation \ref{photo-loss} and is used to calculate optical flow. MotionNet produces optical flow using three novel unsupervised loss functions which we describe below.

A \textbf{pixelwise reconstruction error function} is used to measure the difference in pixel values between the frames:

\begin{equation}
L_{pixel} = \frac{1}{hw} \sum_i^h \sum_j^w \rho(\bm{I}_1(i, j) - \bm{I}_2(i + \bm{V}_{i, j}^x, j + \bm{V}_{i, j}^y))
\label{equ:pixelreconstruction}
\end{equation}

\noindent where $h$ and $w$ are the height and width of images $\bm{I}_1$ and $\bm{I}_2$. The horizontal and vertical components of the estimated optical flow are denoted as $\bm{V}^x$ and $\bm{V}^y$. The term $\rho(x) = (x^2 + \epsilon^2)^{\alpha}$ denotes the generalized Charbonnier loss function which is a differentiable version of the $L_1$ norm used to decrease the influence of outliers, where $\epsilon$ determines how closely it resembles the $L_1$ norm.  The arbitrary power $\alpha$ is non-convex when $\alpha < 0.5$. 

A \textbf{smoothness loss function} resolves ambiguities that occur when estimating motion in non-textured local regions (aperture problem). It is difficult to detect motion on local image patches using convolution if patches are not textured. Let $\nabla \bm{V}_x^y$ and $\nabla \bm{V}_y^y$ be the  gradients of the vertical component of the estimated flow field, and $\nabla \bm{V}_x^x$ and $\nabla \bm{V}_y^x$  be the gradients of the horizontal component. Smoothness loss is defined as:

\begin{equation}
L_{smooth} = \rho (\nabla V_x^x) + \rho (\nabla V_y^x) + \rho (\nabla V_x^y) + \rho (\nabla V_y^y)
\label{equ:smooth}
\end{equation}
 
The \textbf{structural similarity loss function} helps the network learn frame structure. A sliding window approach is used to partition $\bm{I}_1$ and $\bm{I}_1'$ into local patches for comparison.  Let $N$ denote the number of patches that can be extracted from the image using a sliding window with stride $8$, let $n$ denote the patch index, and let $I_{1n}$ and $I'_{1n}$ denote two image patches from the original image $I_1$ and its reconstruction $I'_1$:

\begin{equation}
L_{ssim} = \frac{1}{N} \sum_n^N (1 - SSIM(\bm{I}_{1n}, \bm{I}'_{1n})),
\end{equation}

\noindent where SSIM measures the perceptual quality of frames computed as:

\begin{equation} \label{equ:ssim}
SSIM(\bm{I}_{p1}, \bm{I}_{p2}) = \frac{(2 \mu_{p1} \mu_{p2} + c_1)(2 \sigma_{p1p2} + c_2)}{(\mu^2_{p1} + \mu^2_{p2} + c_1)(\sigma^2_{p1} + \sigma^2_{p2} + c_2)}.
\end{equation}

\noindent SSIM determines the average dissimilarity between images by examining the means $\mu_{p1}$ and $\mu_{p2}$,  variances $\sigma_{p1}$ and $\sigma_{p2}$, and covariance $\sigma_{p1p2}$ of the pairs of image patches. Two constants $c_1$, and $c_2$ are used to stabilize the division.

Loss was computed at each scale $s$ as a weighted sum of the three loss functions, where $\lambda_1$, $\lambda_2$, and $\lambda_3$ weigh the importance of each loss function: 

\begin{equation}
L_{s} = \lambda_1 L_{pixel} + \lambda_2 L_{smooth} + \lambda_3 L_{ssim}
\end{equation}

MotionNet was combined with a temporal stream CNN which learned to map the generated optical flow to class labels for action recognition.  Two methods for combining models were explored. Stacking simply concatenated the two networks with MotionNet placed in front. Branching used a single network for optical flow extraction and action classification, and then shared features between the two tasks.  Stacking proved more space efficient than branching. Multiple strategies were explored for fine-tuning the stacked CNN for action recognition. The strategy with the highest performance fine-tuned both networks (MotionNet and the temporal CNN) utilizing all three loss functions. Long-term motion dependencies were represented using a sequence of $11$ frames which yielded multiple ($10$) optical flow fields. 

The proposed method~\cite{Zhu2019} outperformed real-time action recognition methods at the time (~\cite{Kantorov2014,Zhang2016,Tran2015,Ng2018,Diba2016,Wang2016})  on four benchmarks datasets; UCF101, HMDB51, THUMOS14~\cite{Jiang2014}, and ActivityNet~\cite{Heilbron2015}. THUMOS14 and ActivityNet are large datasets of long, untrimmed videos. MotionNet was shown to be flexible as it may be concatenated to deeper CNN architectures to improve their recognition accuracy. Such frameworks are still real-time due to the on-the-fly optical flow calculation.

\textbf{~\hbox{3D} CNNs} use spatio-temporal convolutions that model both spatial structure and temporal information in sequences. They operate on the premise that~\hbox{3D} convolutions preserve temporal information from input signals while~\hbox{2D} convolutions collapse temporal information.

\begin{figure}
\begin{center}
\includegraphics[width=\linewidth]{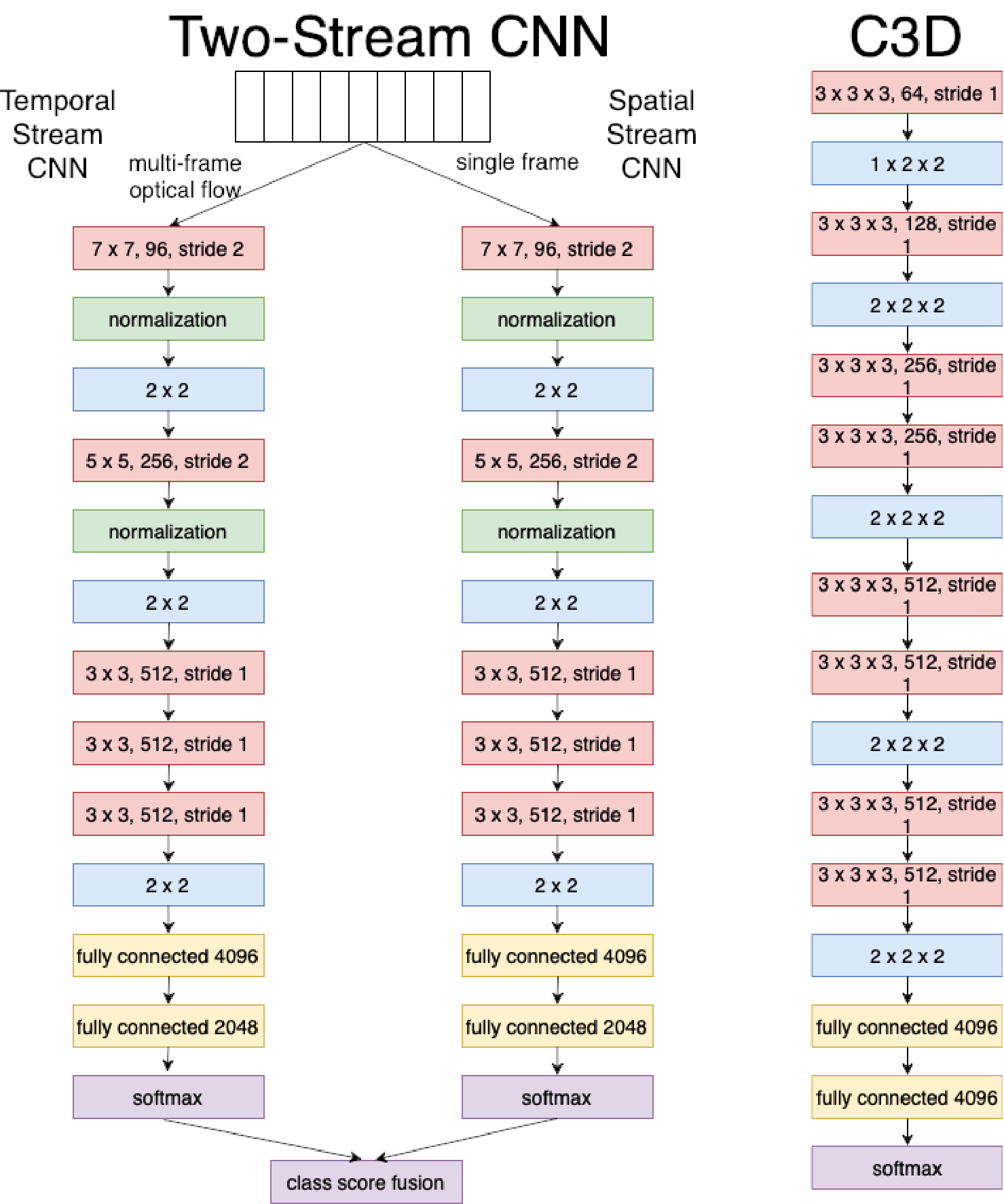}
\caption{Two-Stream Video Classification Architecture (left)~\cite{Simonyan2014} and C3D Architecture (right)~\cite{Tran2015}. Red, green, blue, and yellow layers denote convolution, normalization, pooling, and fully connected layers, respectively. Zoom in to see details. Adapted from~\cite{Simonyan2014} and~\cite{Tran2015}.} \label{fig:twostream-fig}
\end{center}
\end{figure}

C3D~\cite{Tran2015} is a~\hbox{3D} CNN that achieved a new state-of-the-art in object recognition and scene classification in videos. The approach found that the homogenous setting of $3\times 3 \times 3$ kernels in all convolutional layers is the best option for \hbox{3D} CNNs which is consistent with VGGNet (Figure \ref{fig:twostream-fig}, right). The architecture has $8$ convolutional layers with $3 \times 3 \times 3$ convolutional filters, $5$ max pooling layers and $2$ fully connected layers followed by a softmax output layer. Pooling layers are $2 \times 2 \times 2$, except for the first layer which is $1 \times 2 \times 2$ to help preserve temporal information. The stride is $1$ for pooling and convolution. Video clips each consisting of $16$ frames were the network inputs. The architecture was designed to be as deep as possible while accounting for training on a large dataset, GPU memory, and computation affordability.

The C3D network is trained from scratch on the Sports-1M dataset. The trained network was used as a feature extractor for different tasks including action recognition on UCF-101, action similarity labeling, scene and object recognition. The learned features were evaluated for action recognition and showed $5\%$ improvement compared to slow fusion~\cite{Karpathy2014} . Evaluation on UCF101 produced an $11\%$ improvement over the slow fusion model~\cite{Karpathy2014} for action recognition.

Three-dimensional CNNs like C3D have several advantages. The performance of C3D was similar to the original two-stream approach \cite{Simonyan2014}, but with a lower runtime.  C3D does not use optical flow as input like the two-stream architecture~\cite{Simonyan2014} does, and instead extracts motion features directly from videos. There are still some drawbacks. Three-dimensional CNNs are harder to train than \hbox{2D} CNNs because they have many more parameters from the extra filter dimension. In addition, C3D did not benefit from performance improvements from ImageNet pre-training~\cite{Deng2009} because it is a relatively shallow architecture trained from scratch.

A novel two-stream \emph{Inflated}~\cite{Kay2017}~\hbox{3D} CNN, I3D~\cite{Carreira2017}, improved upon the previous state-of-the-art in action recognition~\cite{Simonyan2014,Feichtenhofer2016,Tran2015} by inflating~\hbox{2D} CNN image classification models into~\hbox{3D} CNNs. This inflation process adds a temporal dimension to the filters and pooling kernels in a~\hbox{2D} CNN.

\textbf{Transfer Learning on Video Datasets:} Originally, it was unclear whether transfer learning was as useful for videos. A new dataset, the Kinetics-400 Human Action Video Dataset~\cite{Kay2017}, was collected to evaluate this. This dataset  contains about $306,245$ videos that correspond to $400$ human action classes, with over $400$ clips for each class. The dataset is split into training, test and validation sets. Several architectures including the original two-stream approach~\cite{Simonyan2014}, C3D~\cite{Tran2015}, the~\hbox{3D} fused two-stream approach~\cite{Feichtenhofer2016}, and I3D~\cite{Carreira2017}, were compared by first pre-training on the Kinetics-400 dataset, then fine-tuning on HMDB51 and UCF101. The difference between the performances of the best deep network, I3D~\cite{Carreira2017}, and the best hand-crafted approach, iDT~\cite{Wang2013}, for video action recognition was smaller than the difference between that of the best deep learning-based and hand-crafted image recognition approaches. Additionally, a~\hbox{2D} CNN applied to individual frames of a video performed comparably to I3D on the Sports-1M dataset. This work~\cite{Carreira2017} demonstrated that there is considerable benefit to transfer learning in videos. Kinetics-600~\cite{ Carreira2018}, the second iteration of the Kinetics dataset contains about $500,000$ $10$-second long videos sourced from YouTube that correspond to $600$ classes.

The aforementioned transfer learning results inspired a subsequent work~\cite{Tran2018} that reconsidered~\hbox{3D} CNNs in the context of residual learning. ResNets use skip connections to speed up network learning and permit substantially deeper networks that can learn more complex functions. \hbox{3D} ResNets have additional dimensions, \hbox{3D} convolutions and \hbox{3D} pooling. Three-dimensional ResNets outperformed~\hbox{2D} ResNets in action recognition. The team also investigated several forms of spatio-temporal convolution:~\hbox{2D} convolution on frames,~\hbox{2D} convolution on clips,~\hbox{3D} convolution, \hbox{2D-3D}  convolution, and a factorization of~\hbox{3D}  convolution into~\hbox{2D} spatial convolution followed by~\hbox{1D} temporal convolution ($R(2+1)D$ convolution). Using $R(2+1)D$ convolution doubled the number of nonlinearities in the network. This facilitated gradient optimization and enabled more complex function representations. During evaluation experiments, the $R(2+1)D$ convolutional block achieved state-of-the-art accuracy on the Kinetics-400 and Sports-1M datasets. The team also investigated whether the model supported transfer learning by pre-training on Kinetics-400 and Sports-1M and then fine-tuning on UCF101 and HMDB51. In the transfer learning experiment $R(2+1)D$ outperformed all other methods except for I3D, which uses Kinetics-400 and ImageNet for pre-training. Table \ref{tab:action-recognition-summary-table} summarizes the action recognition approaches discussed in this section, and their performance on UCF101.

\noindent \textbf{Improving Computational Cost of \hbox{3D} CNNs:}
MF-Net uses an ensemble of lightweight networks (fibers) instead of one complex network to improve performance while reducing computation cost. The \hbox{3D} Multi-Fiber architecture was designed for human action recognition and achieved better accuracy on Kinetics-400, UCF101, and HMDB51 compared to other \hbox{3D} CNNs like $I3D$~\cite{Carreira2017} and $R(2+1)D$~\cite{Tran2018} while using fewer FLOPs.

\section{Animal Behavior Classification} \label{sec:animalbehavior}

\begin{figure}[ht]
\begin{center}
\includegraphics[width=\linewidth]{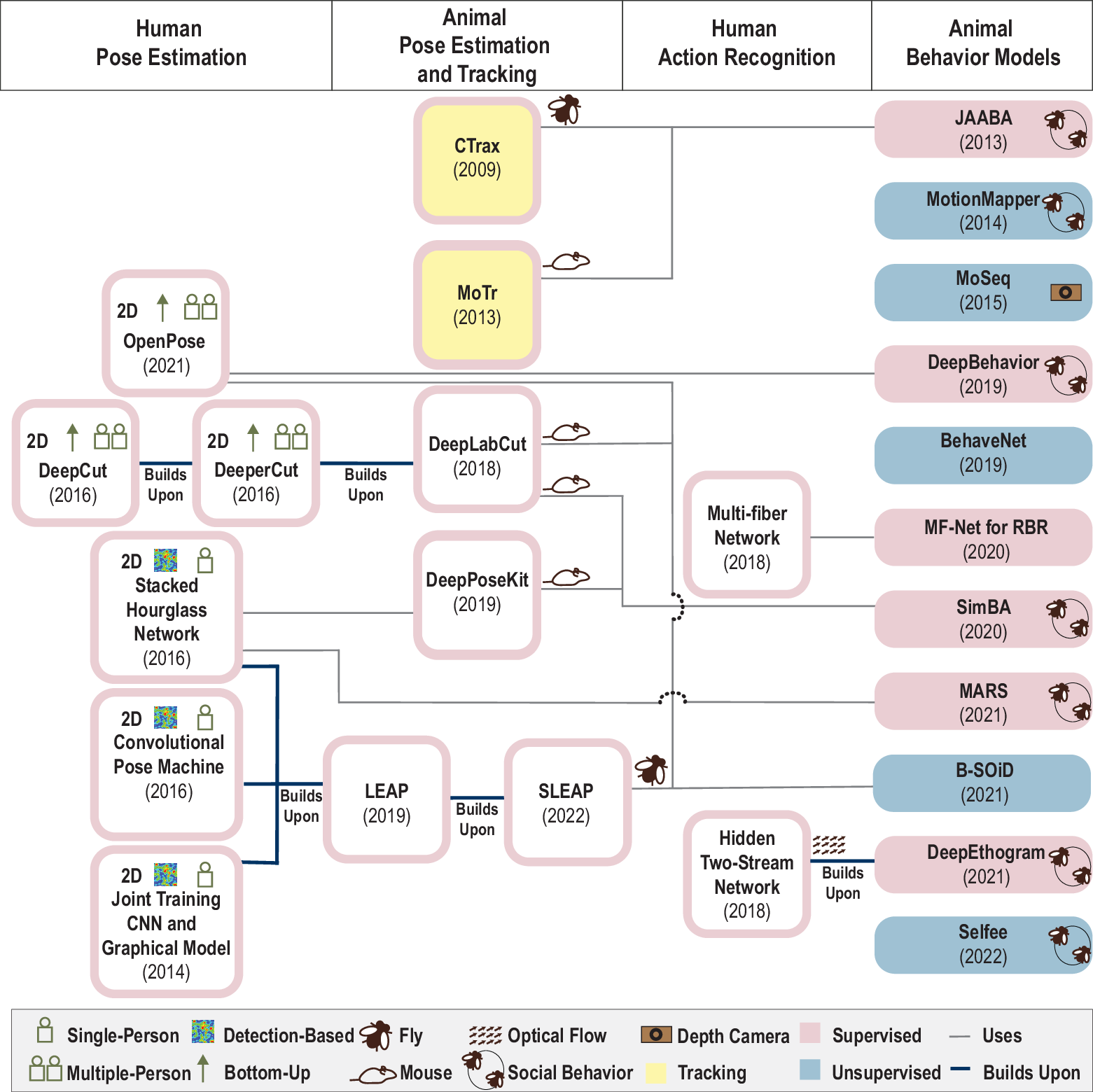}
\caption{Animal Behavior Classification Frameworks and the Action Recognition, Pose Estimation and Tracking Methods they Use or Build Upon.} 
\label{fig:animalbehaviormodelaction}
\end{center}
\end{figure}

This section reviews adaptations of the learning models presented in  Sections~\ref{sec:hpe} and~\ref{sec:actionrecognition} for animal behavior classification. We focus on models that analyze small laboratory animals in neuroscience. Figure~\ref{fig:venndiagram} gives a graphical overview of the classification frameworks, pose estimation and tracking methods organized by the CNN networks they use. Figure~\ref{fig:actionrecognition} shows the relationship of several animal classification approaches to spatio-temporal infrastructures for human action recognition. Figure~\ref{fig:animalbehaviormodelaction} shows the relationship of techniques reviewed in this section to human action recognition and human pose estimation methods. We include  systems that are significantly different so that important developments in the underlying deep learning architectures are represented. We also indicate validation animal types, social interaction support, and whether depth cameras are required. Optical flow methods are identified. Human pose estimation taxonomy categories~\cite{Zheng2020} (Section~\ref{sec:hpe}) are shown.

We begin with background information on manual scoring for labeling videos, with a brief overview of the challenges with traditional vision approaches that lead to breakthroughs in deep learning-based classification solutions. In Sections~\ref{sec:ape2d} and~\ref{sec:ape3d}, we review animal pose estimation as many frameworks require a pose estimation before presenting our taxonomy for organizing these frameworks in Section~\ref{sec:taxonomy}.

\textbf{Manual Behavior Scoring:} Quantification of animal behavior is important in ethology and behavioral neuroscience~\cite{gulinello2019rigor}. However, it is well known that manual scoring of animal behavior videos is time-intensive, costly, and subjective~\cite{gulinello2019rigor}. Subtle fine-level behaviors may go undetected as behaviors must be pre-defined~\cite{reynolds2013effects}. Manual behavioral scoring is also subject to human bias as results can vary between different research labs and between researchers within the same lab~\cite{gulinello2019rigor, Bohnslav2021,Segalin2021}. Moreover, individual scorers themselves may drift over time as they gain experience~\cite{koncz2017measuring}. Thus, scoring by the same individual may change across different sessions.

\textbf{Traditional Methods} use automated software to facilitate manual behavior scoring. There are several challenges with the application  of these methods across research labs. Early automated methods often relied on specialized equipment like depth cameras~\cite{Wiltschko2015} which may be costly or require expertise to operate. Some methods require that markers are placed on animals~\cite{Mathis2018} for motion tracking. The use of markers is less desirable because they can be intrusive, and their locations must be determined a priori. Markers can also distract the animal in a way that alters its behavior and skews experimental results. It is common for traditional systems to use older computer vision processing techniques~\cite{Datta2019} without incorporating neural networks. These methods require supervision in the form of pre-selected labels specified by the researcher. They also require highly controlled laboratory environments and have limited ability to track multiple animals.  Commercial software and other solutions that do not offer publicly available open-source code are harder to reuse and adapt for different experiments.  

\textbf{The Promise of Deep Learning Solutions:} Deep learning methods are better suited for these problems~\cite{Graving2019,Nath2019} as they permit multiple unmarked animals to be tracked in almost any context, including in-the-wild. These techniques are widely applicable because they generally operate using RGB cameras and are often developed as open-source platforms. Some open-source solutions outperform commercial software with added flexibility for customization which can facilitate adoption. Pose-based approaches are relatively subjective and may not generalize across labs and experiments if animals vary in size or camera frame rates are different~\cite{Hsu2021}. Approaches that classify behavior directly from raw pixel values without estimating pose mitigate this issue and simplify the behavior classification pipeline~\cite{Bohnslav2021}. Data dependent factors make systems difficult to generalize pipelines(frame-rate, intensities)~\cite{ Berman2018}. We now review several deep learning algorithms that have led to advancements in behavior analysis in neuroscience.

\subsection{\hbox{2D} Animal Pose Estimation}\label{sec:ape2d}

This section reviews CNN network architectures for \hbox{2D} animal pose estimation that leverage methods discussed in Sections~\ref{sec:hpe} and~\ref{sec:actionrecognition}. We cover \hbox{3D} techniques in the next section. Table~\ref{tab:table-animal-pose-estimation} summarizes  the key contributions of \hbox{2D} and \hbox{3D}  animal pose estimation methods, indicating validation animal types, method type, and the number of cameras required. Coarse level supervision is indicated (pink: supervised, blue: unsupervised). Common animal behavior video datasets are found in Table~\ref{tab:behavior-video-datasets-table}. 

An early deep learning-based approach for \hbox{2D} animal pose~\cite{Wiltschko2015} classified sub-second structure in mice behavior using unsupervised learning. The approach first compressed videos of freely moving mice using principal component analysis (PCA). It then used an auto regressive hidden Markov model to segment the compressed representation into discrete behavioral syllables, identifiable modules (like grooming and rearing) whose sequence is governed by definable transition statistics. The technique requires special depth cameras. These are limiting because they may not detect very small laboratory animals and reconstructing full \hbox{3D} poses is challenging without multiple camera viewpoints.

DeepLabCut~\cite{Mathis2018} was the first tool to extend human pose estimation techniques to the animal pose estimation problem~\cite{MATHIS20201}. The approach modified the feature detector architecture in DeeperCut~\cite{Insafutdinov2016}  which builds upon ResNet (Section~\ref{hpi-from-images} and Figure~\ref{fig:venndiagram}).  The toolkit tracks multiple animals' \hbox{3D} poses in \hbox{2D} images. Camera calibration with one or more cameras is used to find \hbox{3D} keypoint locations. DeepLabCut requires $200$ labeled frames because it uses a model pretrained on ImageNet. Multi-animal DeepLabCut~\cite{Lauer2022} extends DeepLabCut to the multi-animal setting. The authors achieve this by adopting aspects of three CNN architectures~\cite{He2016,Tan2019,Cao2021} and using a transformer~\cite{Zhu2020,Li2021}  architecture to reidentify animals when they leave the field of view.

\begin{table} [h]
\begin{center}
\begin{tabular}{| m{0.25\columnwidth} | m{0.05\columnwidth} | m{0.12\columnwidth} | m{0.07\columnwidth} | m{0.12\columnwidth} | m{0.12\columnwidth} |}
\hline
Dataset & Year & Trimmed & Size & Social & Classes \\
\hline
Clipped Mouse \cite{Jhuang2010} & $2010$ & \cmark & $4,268$ & \xmark & $8$ \\
\hline
Full Mouse \cite{Jhuang2010} & $2010$ & \xmark & $12$ & \xmark & $8$ \\
\hline
CRIM13 \cite{BurgosArtizzu2012} & $2012$ & \xmark & $474$ & \cmark & $13$ \\
\hline
Sturman -- EPM \cite{Sturman2020} & $2020$ & \xmark & $24$ & \xmark & $6$ \\
\hline
Sturman -- FST \cite{Sturman2020} & $2020$ & \xmark & $29$ & \xmark & $2$ \\
\hline
Sturman -- OFT \cite{Sturman2020} & $2020$ & \xmark & $20$ & \xmark & $4$ \\
\hline
Mouse -- \hbox{Ventral1} \cite{Bohnslav2021} & $2021$ & \xmark & $14$ & \xmark & $28$ \\
\hline
Mouse -- \hbox{Ventral2} \cite{Bohnslav2021} & $2021$ & \xmark & $14$ & \xmark & $16$ \\
\hline
Mouse --\hbox{OpenField} \cite{Bohnslav2021} & $2021$ & \xmark & $20$ & \xmark & $6$ \\
\hline
\hbox{Mouse --} HomeCage \cite{Bohnslav2021} & $2021$ & \xmark & $11$ & \xmark & $8$ \\
\hline
Mouse -- Social \cite{Bohnslav2021} & $2021$ & \xmark & $12$ & \cmark & $6$ \\
\hline
Fly \cite{Bohnslav2021} & $2021$ & \xmark & $19$ & \xmark & $2$ \\
\hline
\end{tabular}
\end{center}
\vspace{5pt}
\caption{Common Animal Behavior Video Datasets} 
\label{tab:behavior-video-datasets-table}
\end{table}

LEAP~\cite{Pereira2019} was designed to estimate poses in videos of individual mice and fruit flies. It adapts three CNNs originally developed for human pose estimation~\cite{TompsonJ2014,Wei2016,Newell2016} into a fully convolutional architecture that predicts heat-maps for body parts from frames. The solution uses an active learning (Section~\ref{sec:background}) approach that requires at least $10$ labeled frames. Two-dimensional pose is estimated at each time step. A spectrogram is computed for the time-series of each joint using a continuous wavelet transform. Spectrograms are then concatenated and embedded in a~\hbox{2D} manifold (behavior space) where each video is represented as a feature vector that captures \hbox{2D} pose dynamics across time. The behavior space distribution is then clustered to identify classes of behaviors.  The complete pipeline from pose estimation to behavioral clustering for behavior recognition is semi-supervised. The output of LEAP was used for unsupervised analysis of the fly behavior repertoire.

SLEAP (Social LEAP)~\cite{Pereira2022}, the successor to LEAP, is a general framework for multi-animal pose estimation. It implements bottom-up and top-down pipelines, identity tracking using motion and appearance models, and over $30$ neural network architectures. Our graphical overview (Figure~\ref{fig:venndiagram}) uses the default architecture used in SLEAP, a version of UNet~\cite{Ronneberger2015}. UNet is an encoder-decoder architecture designed for segmentation of biomedical images. SLEAP is designed to be flexible to allow the comparison of part grouping approaches, identity tracking approaches, and neural network architecture properties. SLEAP achieves comparable or improved accuracy when compared to DeepPoseKit, LEAP, and DeepLabCut for single-animal pose estimation with faster inference speed.

DeepPoseKit~\cite{Graving2019} is a single-animal \hbox{2D} pose estimation method. The project evaluates two models: a new stacked DenseNet CNN~\cite{Jegou2017} network, and a modified stacked hourglass network~\cite{Newell2016} that has origins in human pose estimation. In DenseNets~\cite{Huang2017dense}, each layer is connected to every other layer in a feed-forward fashion to mitigate the vanishing gradient problem and decrease parameterization. The stacked hourglass architecture~\cite{Newell2016} is an encoder-decoder network with several pooling and upsampling layers. Each hourglass module processes features to a low resolution then upscales and combines the features from multiple scales. This design captures information at every scale and allows intermediate supervision. Of the two methods, the stacked DenseNet was found to perform better.

Previous pose estimation methods like DeepLabCut have not been effective in tracking multiple identical animals. AlphaTracker~\cite{Chen2020} is a recently developed tool for multiple-animal tracking that adapts a human pose estimation algorithm, AlphaPose, for animal pose estimation. The positions of animals in each frame are detected using YOLOv3~\cite{Redmon2018arxiv}. The resulting bounding boxes are cropped and fed to a Squeeze-and-Excitation network (SE-Net)~\cite{Hu2018} to estimate pose, and the identities of mice are tracked across frames. Squeeze-and-Excitation (SE) networks~\cite{Hu2018} use SE blocks to improve a CNNs representational capacity by explicitly modeling the interdependencies between convolutional feature channels. Training requires labels, though features extracted from the keypoints are clustered into individual or social behaviors using hierarchical clustering in an unsupervised manner. Calibration is automated. 

\vspace{10pt}
\noindent\textbf{Speed Accuracy Trade-Off}
When choosing a pose estimation architecture, the speed-accuracy trade-off must be considered~\cite{Huang2017speed}. DeepLabCut prioritizes accuracy over speed by using a large over-parameterized model (ResNet). This overparameterization leads to relatively slow inference but maintains accurate predictions. The LEAP system \cite{Pereira2019} prioritizes speed by using a smaller less-robust model. Although it has a faster inference time, it is less robust to data variance and less likely to generalize to different environments. DeepPoseKit aims to improve processing speed without sacrificing accuracy, and reports that the approach is about three times as accurate as LEAP, as accurate as DeepLabCut~\cite{Mathis2018}, and twice as fast as both (Table \ref{tab:poseaccuracy}). 

\begin{table}[h] 
\begin{center}
\begin{tabular}{|c | c | c|} 
 \hline
 Approach & Mean Accuracy & Inference Speed (Hz) \\
 \hline
 DeepLabCut~\cite{Mathis2018} & $0.33$ & $520$ \\
 \hline
 LEAP~\cite{Pereira2019} & $0.13$ & $560$ \\
 \hline
 DeepPoseKit~\cite{Graving2019} & $0.35$ & $1000$ \\
\hline
\end{tabular}
\end{center}
\vspace{5pt}
\caption{Inference Speed and Posterior Mean Accuracy for \hbox{2D} Animal Pose Estimation Approaches. Data sourced from~\cite{Graving2019}.}
\label{tab:poseaccuracy}
\end{table}

\subsection{\hbox{3D} Animal Pose Estimation}\label{sec:ape3d}
A two-dimensional pose is an incomplete representation of an animal's behavior as occlusions (that hide limbs) result in loss of important information. Camera viewing parameters also affect how movement is quantified. DeepLabCut used a printed checkerboard target to calibrate cameras for~\hbox{3D} pose estimation. This type of target would not work well for analyzing small animals like flies (the Drosophila melanogaster fly is $2.5$ mm long). Calibration patterns would have to be extremely small, and color would introduce imaging artifacts at translucent body parts. DeepFly3D~\cite{Gunel2019} adapted the stacked hourglass network architecture~\cite{Newell2016} and added tools to compute the accurate~\hbox{3D} pose of Drosophila melanogaster using the fly itself as the camera calibration target.  Although the approach can be used for any animal type or species, it requires multiple camera views. Like LEAP,~\hbox{2D} pose data and~\hbox{3D} joint angles are used for unsupervised behavioral classification by clustering after~\hbox{3D} pose estimation. When the two data sources were compared, classification using \hbox{3D} joint angles resulted in clearer clusters of behaviors. LiftPose3D~\cite{Gosztolai2021} only requires one camera view, a prism mirror, and $1,000 - 10,000$ training images.  The approach is based on a CNN architecture developed for mapping~\hbox{2D} human poses to~\hbox{3D} poses using a single camera view~\cite{Martinez2017} (Section ~\ref{sec:hpe}). Another method, Anipose~\cite{Karashchuk2021}, built on DeepLabCut, allows \hbox{3D} reconstruction from multiple cameras using a wider variety of methods. 

MouseVenue3D~\cite{Han2022} contributed an automated \hbox{3D} behavioral capture system that uses multi-view cameras to estimate \hbox{3D} pose in marker-less rodents. This work showed that \hbox{3D} behavioral data yields improved performance in behavior recognition tasks compared to \hbox{2D} behavioral data.

\subsection{Tracking} \label{sec:tracking}

Here, we review two tracking methods used by the animal behavior classification frameworks we discuss.

CTrax~\cite{Branson2009} is a computer vision system designed to identify the position and orientation of multiple flies in a 2D arena in each frame of input videos. Flies are detected first through background subtraction and then fitting an ellipse to each fly by fitting a Gaussian to each fly's pixel locations. Next, flies are assigned identities using the Hungarian algorithm for minimum-weight perfect bipartite matching.

MoTr (Mouse Tracker)~\cite{Ohayon2013} is a computer vision system similar to CTrax~\cite{Branson2009} designed for tracking multiple mice in a 2D arena. Flies are detected through background subtraction and fitting ellipses to mice using the expectation-maximization algorithm. Then a hidden Markov model is used to associate mouse identities with trajectories. Multiple mice are recognized in overhead videos by applying distinct patterns to the back of each mouse and training a classifier on histogram of oriented gradients (HOG)~\cite{Dalal2005} features extracted from image patches containing individual mice.

\subsection{Taxonomy for Animal Behavior Classification} \label{sec:taxonomy}

This section provides a high-level overview of animal behavior classification frameworks for small laboratory animals. We outline a general taxonomy that organizes methods as supervised or unsupervised at a coarse level, and with varying degrees of supervision influenced by inputs, framework components and learning strategies. Supervised methods require manual labels for classification while unsupervised methods automatically cluster the data and assign labels. Within these groups we consider the influence of hand-crafted features, labels (Pose or Behavior), pose estimation (PE), optical flow, and residual learning. We focus on optical flow and residual learning as a study~\cite{Nguyen2019}  showed improvement when using \hbox{3D} CNNs pre-trained to recognize human behaviors, then fine-tuned on mouse behavior datasets.  This study used two-stream fusion with I3D and R(2+1)D by calculating optical flow on a dataset of solitary mice videos~\cite{Jhuang2010}. R(2+1)D uses residual learning. Table~\ref{tab:table-animal-action} summarizes key contributions of these frameworks with coarse level of supervision indicated (pink: supervised, blue: unsupervised). The table also indicates whether pose estimation or depth cameras are required, the type of animal used for validation, and whether there is support for individual or social interactions. Figures~\ref{fig:posebased},~\ref{fig:SNoPE} and ~\ref{fig:USNoPE} diagram a representative set of systems (posed-based, supervised without pose estimation, and unsupervised without pose estimation respectively).

\begin{figure*}
\begin{center}
\begin{tabular}{ |l|l| }
\hline
\multicolumn{2}{ |c| }{Taxonomy for Animal Behavior Classification} \\
\hline
\rowcolor{lightPink}
\multicolumn{2}{ |c| }{Supervised Classification} \\
\hline
\rowcolor{lightPink}
 Hand-crafted Features, Behavior Labels & \cite{Rousseau2000}, \cite{Dankert2009}, \cite{Jhuang2010}, \cite{BurgosArtizzu2012}, \cite{Giancardo2013}, \cite{Kabra2013}, \cite{Hong2015}, \cite{Lorbach2018}, \cite{DeChaumont2019}, \cite{Gabriel2022} \\
 \hline
\rowcolor{lightPink}
Behavior Labels & \cite{Le2019}, \cite{VanDam2020}, \cite{Hu2022}\\
\hline
\rowcolor{lightPink}
\hbox{Hand-crafted Features, Pose and Behavior Labels, PE} & \cite{Segalin2021}, \cite{Stewart2016}, \cite{Sturman2020}, \cite{Arac2019}, \cite{Li2022}, \cite{Nilsson2020} \\
\hline
\rowcolor{lightPink}
Pose and Behavior Labels, PE & \cite{Zhou2022b} \\
\hline
\rowcolor{lightPink}
Optical Flow, Hand-crafted Features, Behavior Labels & \cite{VanDam2013}\\
\hline
\rowcolor{lightPink}
\hbox{Residual Learning, Optical Flow, Behavior Labels} & \cite{Bohnslav2021}\\
\hline
\rowcolor{lightPink}
\hbox{Residual Learning, Pose and Behavior Labels, PE} & \cite{Zhou2022}\\
\hline
\rowcolor{lightPink}
Residual Learning, Optical Flow, Behavior Labels & \cite{Marks2022}\\
\hline
\rowcolor{lightBlue}
\multicolumn{2}{ |c| }{Unsupervised Classification} \\
\hline
\rowcolor{lightBlue}
Hand-crafted Features, Pose Labels, PE & \cite{Hsu2021}\\
\hline
\rowcolor{lightBlue}
Fully Unsupervised & \cite{Stephens2008}, \cite{Berman2014}, \cite {Wiltschko2015}, \cite{Batty2019}, \cite{Brattoli2021}, \cite{Jia2022}\\
\hline
\end{tabular}
\caption{Taxonomy for Animal Behavior Classification.Pink: supervised, Blue: unsupervised}
\label{tab:taxonomy}
\end{center}
\end{figure*}

\vspace{10pt}
\noindent\textbf{Supervised Classification} requires behavior labels.

       \begin{itemize}
\vspace{6pt}
\item \textbf{Hand-crafted Features, Behavior Labels:} These examples use computer vision processing methods to fit shapes (ellipses) to the animal body, extract features, and apply neural networks to classify learned features to behaviors. 
		\begin{description}
\vspace{6pt}
\item [\textbf{--}]One of the first works to apply neural networks to animal behavior analysis~\cite{Rousseau2000} trained a three-layer feed-forward neural network on $3$ parameters that describe a rat's position, achieving a $63\%$ accuracy over nine behaviors.

\vspace{6pt}
\item [--] Caltech Automated Drosophila Aggression-Courtship Behavioral Repertoire Analysis (CADABRA)~\cite{Dankert2009} measured social behaviors from overhead videos of fly pairs. Fly bodies were localized~\cite{Bishop2006}, and fitted with ellipses before calculating features (size, position and velocity). K-nearest-neighbors was used to classify actions from the features. 

\vspace{6pt}
\item [\textbf{--}]A later study~\cite{Jhuang2010} extracted per-frame features of a rodent and used a SVM hidden Markov model (SVMHMM)~\cite{Altun2003} to output behavioral labels. The primary dataset was a single black mouse with a white background. A dataset with per-frame labels of eight solitary behaviors of a mouse in a standard enclosure over ten hours was also released. 

\vspace{6pt}
\item [\textbf{--}] The next development~\cite{BurgosArtizzu2012} focused on social interaction between two mice with unconstrained colors. The algorithm tracks rodents, calculates features, and applies AdaBoost to classify trajectory and spatiotemporal context features. The Caltech Resident-Intruder Mouse (CRIM13) dataset \cite{BurgosArtizzu2012} was released  with $13$ action classes and $237$ $10$-minute videos ($8$ million frames, over $88$ hours) of top and side views of mice pairs engaging in social behavior. A $2013$ study~\cite{Giancardo2013} generalized this framework to support more mice and different experimental settings using a temporal random forest classifier.

\vspace{6pt}
\item [\textbf{--}]Janelia Automatic Animal Behavior Annotator (JAABA) system \cite{Kabra2013} (Figure~\ref{fig:SNoPE}) fits ellipses to body outlines using CTrax (Section~\ref{sec:tracking}), and then calculates per-frame features from the trajectory. A Gentleboost algorithm~\cite{Friedman2000} classifies behaviors using per-frame \emph{window} features and manually collected behavior labels.

\vspace{6pt}
\item [\textbf{--}] Later Hong et. al.~\cite{Hong2015} trained a random forest classifier on location, appearance, and movement features extracted from egocentric depth maps, and five learned pose parameters to  identify social rodent behaviors. A social behavior analysis project~\cite{Lorbach2018} used GMM and EM algorithms (like CTrax) to track features.

\vspace{6pt}
\item [\textbf{--}]LiveMouseTracker~\cite{DeChaumont2019} is a method for real-time tracking of rodent trajectories for behavior analysis of up to four mice. An infrared-depth camera segments and removes the background from frames, and a random forest algorithm identifies rodents and their orientations (shape and posture) from labels. Individual and social behavioral traits were extracted.

\vspace{6pt}
\item [\textbf{--}] DEcoding Behavior Based on Positional Tracking (BehaviorDEPOT) \cite{Gabriel2022} is a software program for behavior recognition in videos. It uses heuristics to measure human-defined behaviors from pose estimation data (from DeepLabCut) that are easier to interpret than traditional supervised machine learning methods. 
		\end{description}
	\end{itemize}

       \begin{itemize}
\vspace{6pt}
\item \textbf{Behavior Labels:} These approaches use labels but avoid hand-crafted features as they must be defined a priori and may not generalize to different behaviors~\cite{Hu2022}. 
		\begin{description}

\vspace{6pt}
\item [\textbf{--}] A \hbox{3D} CNN~\cite{Le2019} similar to C3D was used to extract short-term spatio-temporal features from clips. These features are fed to an RNN with LSTM blocks to produce long-term features. A SoftMax layer outputs the probabilities of behaviors in each clip. This approach slightly underperformed when compared to an approach that used handcrafted features~\cite{Jhuang2010}.

\vspace{6pt}
\item [\textbf{--}] A multi-fiber neural network (MF-Net)~\cite{VanDam2020,Chen2018} (Figure~\ref{fig:SNoPE}) has been used to classify videos for rat behavior recognition (RBR). The architecture has one \hbox{3D} convolutional layer followed by four multifiber convolutional blocks. Results outperformed the prior work of this team~\cite{VanDam2013} when tested on the same dataset (see discussion of EthoVision XT RBR). However, MF-Net uses data augmentation which did not generalize to different set-ups or animals. Note one less successful option in the paper used optical flow.

\vspace{6pt}
\item [\textbf{--}]The LabGym~\cite{Hu2022} framework provides customizable combinations of neural networks to assess spatiotemporal information in videos. It runs easily on a common laptop, without the need for powerful GPUs required by DeepEthogram~\cite{Bohnslav2021}  and SIPEC~\cite{Marks2022} (which we describe later). It is not optimized    for social behavior analysis but is effective for invertebrate and mammal behavior analysis.		
	\end{description}
		\end{itemize}

       \begin{itemize}
\vspace{6pt}
\item \textbf{Hand-crafted Features, Pose and Behavior Labels, PE:}
		\begin{description}
\vspace{6pt}
\item [\textbf{--}] The Mouse Action Recognition
System (MARS)~\cite{Segalin2021} (Figure~\ref{fig:posebased}) focuses on social behaviors between a black and white mouse. It adapts a stacked hourglass architecture (Section~\ref{sec:ape2d}) with eight hourglass subunits to estimate pose. The rolling feature-window method from JAABA is used to calculate handcrafted features over a time window. XGBoost~\cite{Chen2016b} is trained to predict behaviors given the window features.

\vspace{6pt}
\item [\textbf{--}]The DeepBehavior toolbox~\cite{Arac2019} (Figure~\ref{fig:posebased}) uses a version of GoogLeNet~\cite{Stewart2016} or OpenPose~\cite{Cao2021}  (depending on the species) for kinematic analysis of three rodent behaviors in~\hbox{3D}. Stewart and colleagues designed an architecture~\cite{Stewart2016} for multiple object detection using GoogLeNet to encode images into high level descriptors before decoding the representation into a set of bounding boxes using an LSTM layer. Experiments were conducted using one or two cameras.

\vspace{6pt}
\item [\textbf{--}]Simple Behavior Analysis (SimBA)~\cite{Nilsson2020} (Figure~\ref{fig:posebased}) classifies behaviors based on the locations of certain key points, or joint locations, obtained from~\hbox{2D} pose estimation methods (like DeepLabCut, LEAP and DeepPoseKit). After detecting poses, outliers are identified and corrected. There are $498$ metrics calculated from corrected pose data over rolling windows (like JAABA). Random forest classifiers are trained to predict behavior classes given the calculated features. In a related approach~\cite{Sturman2020}, DeepLabCut is used for per-frame pose estimation, before applying a feedforward neural network to classify pose features as behaviors. Sturman et al.~\cite{Sturman2020} released three video datasets of mice behavioral tests, including the open field test (OFT), elevated plus maze (EPM), and forced swim test (FST).

\vspace{6pt}
\item [\textbf{--}] OpenLabCluster~\cite{Li2022} contributed unsupervised clustering (Section~\ref{sec:background}) of pose features using a deep recurrent encoder-decoder architecture and an active learning approach (Section~\ref{sec:background}). In the latter approach, at each iteration, one sample is labeled, and the cluster map is refined. A fully-connected network is appended to the encoder to perform classification trained on labeled samples. The results produce fast and accurate classification in four animal behavior datasets with sparse labels.

		\end{description}
	\end{itemize}

       \begin{itemize}
\vspace{6pt}
\item \textbf{Pose and Behavior Labels, PE:}
		\begin{description}
\vspace{6pt}
\item [\textbf{--}] ConstrastivePose~\cite{Zhou2022b} uses contrastive learning to reduce differences in pose estimation features and its random augmented version, while increasing differences with other examples. These features have a similar structure as handcrafted features and perform comparably on semi-supervised behavior classification.
		\end{description}
	\end{itemize}

\begin{itemize}
\vspace{6pt}
\item \textbf{Optical Flow, Hand-crafted Features, Behavior Labels:}
       \begin{description}
\vspace{6pt}
\item [\textbf{--}] The EthoVision XT RBR~\cite{ethovisionXT8,VanDam2013} is commercial software that tracks features using optical flow estimated with the Lucas-Kanade algorithm~\cite{Lucas1981,Baker2004}. Features are reduced to a low-dimensional space using the Fisher linear discriminant analysis~\cite{Fukunaga2013} before applying a quadratic discriminant for classification. The testing dataset (solitary rats with per-frame labels of $13$ behaviors) is not public. 
		\end{description}
	\end{itemize}

\begin{itemize}
\vspace{6pt}
\item \textbf{Residual Learning, Optical Flow, Behavior Labels:}
		\begin{description}
\vspace{6pt}
\item [\textbf{--}] DeepEthogram~\cite{Bohnslav2021} (Figure~\ref{fig:SNoPE}) generates an ethogram (a set of user defined behaviors of interest). The system builds upon a two-stream approach that computes optical flow on the fly. First the flow generator is trained on unlabeled videos, while the user labels each frame of some videos in parallel. ResNet models extract per-frame flow and spatial features.  A CNN with Temporal Gaussian Mixture (TGM) layers \cite{Piergiovanni2018}  learns a latent hierarchy of sub-event intervals from untrimmed videos for temporal action detection. The TGM has a large temporal receptive field, and outputs the final probabilities that indicate whether the behavior exists at each frame. Thus, this method integrates aspects of action detection methods (Section~\ref{sec:motivation}) as information from seconds ago helps classify current behaviors. Five benchmark datasets of mice in different arenas were released, including one with social interaction, and one with flies to test if the system generalized to other species. 
		\end{description}
	\end{itemize}

       \begin{itemize}
\vspace{6pt}
\item \textbf{Residual Learning, Pose and Behavior Labels, PE:}
		\begin{description}
\vspace{6pt}
\item [\textbf{--}] A recent work~\cite{Zhou2022} models interactions between mice using a Cross-Skeleton Interaction Graph Aggregation Network (CS-IGANet). A Cross-Skeleton Node-level Interaction (CS-NLI) models multi-level interactions between mice and fuses multi-order features. An Interaction-Aware Transformer for the dynamic graph updates, and node-level representations, was proposed. DeepLabCut is used to estimate pose for datasets without pose labels. A self-supervised task for measuring the similarity of cross-skeleton nodes was also proposed. This approach outperforms other approaches on the CRIM13 dataset.
		\end{description}
	\end{itemize}

       \begin{itemize}
\vspace{6pt}
\item \textbf{Residual Learning, Optical Flow, Behavior Labels:}
		\begin{description}
\vspace{6pt}
\item [\textbf{--}]Segmentation, Identification, Pose-Estimation and Classification (SIPEC)~\cite{Marks2022} contributed multiple deep neural network architectures for individual and social animal behavior analysis in complex environments. These networks operate on pixels. SegNet~\cite{Badrinarayanan2017} is a well-known CNN architecture for pixel-wise segmentation of images on which SIPEC:SegNet is built. SIPEC:IdNet uses DenseNet to produce visual features that are integrated over time through a gated recurrent unit network~\cite{Chung2014} to reidentify animals when they enter or exit the scene. SIPEC:PoseNet performs top-down multi-animal pose estimation. SIPEC:BehaveNet uses an Xception network~\cite{Chollet2017} and a Temporal Convolution Network~\cite{Oord2016} to classify behaviors from raw pixels. This system was the first to classify social behaviors in primates without pose estimation. Note that the classification pipeline (which is our focus) is not pose-based.
		\end{description}
	\end{itemize}

\vspace{10pt}
\noindent \textbf{Unsupervised Classification:} These methods do not require behavior labels for the classification step. The learning process may require input from a supervised pose estimation system (that requires labels). However, the final classification is determined by an automated process that generates and assigns labels (like numeric values) to the output of a clustering system.

\begin{itemize}
\vspace{6pt}
\item \textbf{Hand-crafted Features, Pose Labels, PE:}
       \begin{description}
\vspace{6pt}
\item [\textbf{--}] Behavioral Segmentation of Open-field in DeepLabCut (B-SoiD)~\cite{Hsu2021} (Figure~\ref{fig:posebased}), uses DeepLabCut (Section~\ref{sec:ape2d}),  to obtain six joint locations (snout, four paws, tail base) per-frame, then calculates spatiotemporal relationships (speed, angular change, and distance between joints). Features (like speed) are projected from a High-dimensional space to a low-dimensional space using UMAP~\cite{McInnes2018}. HDBScan~\cite{Campello2015} clusters the results, and a random forest classifier is trained to predict behaviors directly from high-dimensional features without the computationally expensive clustering step. The need to evaluate and transfer non-linearities from dimensionality reduction is also avoided.
       	\end{description}
	\end{itemize}

       \begin{itemize}
\vspace{6pt}
\item \textbf{Fully Unsupervised:}
		\begin{description}

\vspace{6pt}
\item [\textbf{--}] PCA has been applied to a worm's centerline (which describes most of a worm's behavior) to represent the \emph{eigenworm}~\cite{Stephens2008} as three values. Unsupervised learning has been used to detect behaviors as sequences of eigenworm positions~\cite{Brown2013}, before clustering using affinity propagation~\cite{Frey2007}.

\vspace{6pt}
\item [\textbf{--}] MotionMapper~\cite{Berman2014} (Figure~\ref{fig:USNoPE}) classifies fruit fly behaviors. The system shows that fruit fly posture can be captured by a low dimensional subspace of eigenmodes. PCA analysis projects a processed (segmented, resized and aligned) frame (to a set of 50-dimensional time-series. A Morlet wavelet transform~\cite{Goupillaud1984}  is then applied to create a spectrogram for each postural mode. The wavelet's multi-resolution time-frequency trade-off reflects postural dynamics at different time scales. Then, t-SNE~\cite{VanDerMaaten2008} is used to embed the spectrograms to a \hbox{2D} space and preserve the local structure of clusters. Finally, a watershed transform~\cite{Meyer1994} segments the embedding and identifies individual behavior peaks. In a related approach~\cite{Vogelstein2014}, a type of hierarchical clustering is used to describe the behaviors of Drosophila larvae. This experiment characterized behavior through the identification of $29$ atomic movements and four movement categories. MotionMapper~\cite{Klibaite2017} was later extended to classify paired fruit-fly behaviors by first segmenting the two flies in each frame.

\vspace{6pt}
\item [\textbf{--}] Motion Sequencing (MoSeq)~\cite{Wiltschko2015, Wiltschko2020}, (Figure~\ref{fig:USNoPE}), embeds a depth video of an animal into a ten-dimensional space using PCA. An AR-HMM is then fit to the principal components to generate behavioral syllables, producing  a dynamical and behavioral representation. The need for a parameter that controls the time scale of transitions between behavioral states can be eliminated by finding motifs of varied lengths~\cite{Brown2013} or using time-frequency analysis like a wavelet transform~\cite{Berman2014,Klibaite2017}. Wavelet transforms can capture multi-scale postural dynamics.

\vspace{6pt}
\item [\textbf{--}] BehaveNet~\cite{Batty2019} (Figure~\ref{fig:USNoPE}) builds upon MoSeq~\cite{Wiltschko2015}. It uses convolutional autoencoders (CAEs) instead of PCA to generate a low-dimensional continuous representation of behavior. An AR-HMM is used to segment the representation into behavioral syllables.

\vspace{6pt}
\item [\textbf{--}] Unsupervised behavior analysis and magnification (uBAM) \cite{Brattoli2021} introduces a neural network designed to compare behaviors, and another designed to magnify differences in behavior. This allows classification behavior extraction and analysis of small behavioral differences, without using any pose or behavior labels.

\vspace{6pt}
\item [\textbf{--}] Self-supervised Feature Extraction (Selfee)~\cite{Jia2022} (Figure~\ref{fig:USNoPE}), trains a network to extract suitable classification features from unlabeled videos (about $5$ million). Two Siamese CNNs based on ResNet-50 learn a discriminative feature representation of behavior with a training rate that is faster ($8$ hours on one GPU) than modern self-supervised models. Note that this paper also included  a classification example that used labels.
		\end{description}
\end{itemize}

\begin{table*}[t]
\begin{center}
\begin{tabular}{| m{2cm} | m{0.5cm} | m{1.3cm} | m{1.8cm} | m{0.75cm} | m{1.2cm} | m{7.5cm} |} 
 \hline
 Method & Year & Multiple Animals & Validated & 2D/3D & Multiple Cameras & Key Contribution(s) \\ 
 \hline
\rowcolor{lightPink}
DeepLabCut \cite{Mathis2018} & $2018$ & \xmark & \hbox{Mice, Flies,} Humans, Horses, Fish & 2D/3D & \cmark &
\begin{itemize}[leftmargin=*] 
            \item The application of the DeeperCut \cite{Insafutdinov2016} feature detector architecture, that was originally designed for human pose estimation, to animal pose estimation.
    \end{itemize}\\
\hline
\rowcolor{lightPink}
LEAP \cite{Pereira2019} & $2019$ & \xmark & Mice, Flies & 2D & \xmark & 
\begin{itemize}[leftmargin=*] 
            \item The design of a CNN architecture inspired by three human pose estimation architectures~\cite{TompsonJ2014,Wei2016, Newell2016}. The CNN is shallow and is trained from scratch to improve inference speed compared to DeepLabCut at the expense of less robustness and requiring more training data to achieve the same performance. 
    \end{itemize}\\ \hline
\rowcolor{lightPink}
DeepPoseKit \cite{Graving2019} & $2019$ & \cmark \hbox{(if not} occluded) & \hbox{Flies, Locusts,} Zebras & 2D & \xmark & 
\begin{itemize}[leftmargin=*] 
            \item The development of a stacked DenseNet architecture \cite{Newell2016} \cite{Jegou2017}~\cite{Huang2017dense} to improve speed and robustness.
    \end{itemize} \\
 \hline
\rowcolor{lightPink}
DeepFly3D \cite{Gunel2019} & $2019$ & \xmark & Flies, \hbox{Humans} & 3D & \cmark & 
\begin{itemize} [leftmargin=*]
            \item The adaptation of a stacked hourglass architecture \cite{Newell2016} to permit 3D pose estimation using multiple cameras.
    \end{itemize} \\
 \hline
\rowcolor{lightPink}
Anipose \cite{Karashchuk2021} & $2021$ & \xmark & \hbox{Flies, Mice,} \hbox{Humans}  & 3D & \cmark & 
\begin{itemize}[leftmargin=*] 
            \item The introduction of filters to improve 2D tracking and a novel triangulation procedure for 3D pose estimation using multiple cameras. 
    \end{itemize}\\
 \hline
\rowcolor{lightPink}
LiftPose3D \cite{Gosztolai2021} & $2021$ & \xmark & \hbox{Flies, Mice,} Monkeys & 3D & \xmark & 
\begin{itemize}[leftmargin=*] 
            \item The adaptation of an architecture designed for lifting 2D human pose into 3D~\cite{Martinez2017} for monocular 3D animal pose estimation.
    \end{itemize}\\
 \hline

\rowcolor{lightPink}
Multi-animal DeepLabCut \cite{Lauer2022} & $2022$ & \cmark & Mice, \hbox{Monkeys,} Fish & 2D/3D & \cmark &
\begin{itemize} [leftmargin=*]
            \item The development of a multi-task architecture which identifies keypoints, assembles them to predict limbs, then assigns them to individuals. This formulation allows for the tracking of multiple animals in one scene.
    \end{itemize} \\
\hline

\rowcolor{lightPink}
SLEAP \cite{Pereira2022} & $2022$ & \cmark & \hbox{Mice, Flies,} Bees, Gerbils & 2D & \xmark & 
\begin{itemize}[leftmargin=*] 
            \item The design of a novel multi-animal pose tracking system that incorporates multiple CNN architectures and approaches for part grouping and identity tracking.
    \end{itemize} \\
 \hline

\end{tabular}
\end{center}
\vspace{5pt}
\caption{Animal Pose Estimation Methods. Pink: supervised.} 
\label{tab:table-animal-pose-estimation}
\end{table*}

\begin{table*}
\begin{center}
\begin{tabular}{| m{2cm} | m{0.5cm} | m{0.7cm} | m{0.7cm} | m{1.2cm} | m{0.7cm} | m{8cm} |} 
 \hline
 Method & Year & Pose Based & Social & Validated & Depth Cam. & Key Contribution(s) \\ 
 \hline
\rowcolor{lightPink}
JAABA \cite{Kabra2013} & $2013$ & \cmark & \cmark & Mice, Flies & \xmark & 
\begin{itemize} [leftmargin=*] 
            \item The development of a supervised behavior classification system usable by biologists.
            \item A set of per-frame features are computed from the pose trajectories that are input to an optimized GentleBoost learning algorithm~\cite{Friedman2000}.
   \end{itemize} \\
 \hline
\rowcolor{lightBlue}
MotionMapper \cite{Berman2014} & $2014$ & \xmark & \cmark & Flies & \xmark &  
\begin{itemize} [leftmargin=*] 
            \item The design of an unsupervised behavior classification pipeline in invertebrates in which frames are segmented, scaled, aligned, then decomposed via PCA. 
            \item A spectrogram is created for each postural mode, then each point in time is mapped to a 2D plane using t-SNE. Peaks are identified after the application of a watershed transform.
   \end{itemize} \\
 \hline
\rowcolor{lightBlue}
MoSeq \cite{Wiltschko2015, Wiltschko2020} & 2015 & \xmark & \xmark & Mice & \cmark &  
\begin{itemize} [leftmargin=*] 
            \item The design of an unsupervised approach to identify behavioral modules in vertebrates.
            \item After compressing mice videos using PCA, an AR-HMM is used to segment the representation into discrete behavioral syllables.
   \end{itemize} \\
 \hline
\rowcolor{lightPink}
DeepBehavior \cite{Arac2019} & $2019$ & \cmark & \cmark & Mice, \hbox{Humans} & \xmark &  
\begin{itemize} [leftmargin=*] 
            \item The adaptation of GoogLeNet~\cite{Szegedy2015} and YOLO-v3 \cite{Redmon2018arxiv} CNN architectures to identify rodent behaviors individually and socially, respectively.
   \end{itemize} \\ 
 \hline
\rowcolor{lightBlue}
BehaveNet \cite{Batty2019} & $2019$ & \xmark & \xmark & Mice & \xmark &  
\begin{itemize} [leftmargin=*] 
            \item The introduction of a probabilistic framework for the unsupervised analysis of behavioral videos and neural activity. 
            \item The use of a CAE to compress videos before they are segmented into discrete behavioral syllables using an AR-HMM.
            \item A generative model that can be used to simulate behavioral videos given neural activity.
   \end{itemize} \\
 \hline
\rowcolor{lightBlue}
B-SoiD \cite{Hsu2021} & $2021$ & \cmark & \xmark & Mice & \xmark &  
\begin{itemize} [leftmargin=*] 
            \item The dimensionality reduction of pose features produced by DeepLabCut \cite{Mathis2018} using UMAP. 
            \item The use of HDBScan to cluster the embedded features and the subsequent training of a random forest classifier to predict behaviors directly from high-dimensional pose features. 
   \end{itemize} \\
 \hline
\rowcolor{lightPink}
SimBA \cite{Nilsson2020} & $2020$ & \cmark & \cmark & Mice, Rats & \xmark &  
\begin{itemize}[leftmargin=*]  
            \item The design of a pose-based approach to rodent social behavior analysis with tools to increase ease-of-use for behavioral neuroscientists.
            \item The classification of behaviors based on positions of recognized joints. 
   \end{itemize} \\
 \hline
\rowcolor{lightPink}
MARS \cite{Segalin2021} & $2021$ & \cmark & \cmark & Mice & \xmark &  
\begin{itemize} [leftmargin=*] 
            \item The development of a pose-based approach to recognize social behaviors between a black and white mouse.
   \end{itemize} \\
 \hline
\rowcolor{lightPink}
DeepEthogram \cite{Bohnslav2021} & $2021$ & \xmark & \cmark & Mice, Flies & \xmark &  
\begin{itemize} [leftmargin=*] 
            \item The design of a novel supervised approach that operates on raw pixel values. 
            \item Optical flow is estimated from video clips using MotionNet \cite{Zhu2019}, flow and spatial features are compressed, and then a sequence model estimates the probability of each behavior in each frame.
   \end{itemize} \\
 \hline
\end{tabular}
\end{center}
\vspace{5pt}
\caption{Animal Behavior Classification Frameworks. Pink: supervised, Blue: unsupervised} \label{tab:table-animal-action}
\end{table*}

\begin{figure*}
\begin{center}
\begin{tabular}{ c }
\includegraphics[width=\textwidth]{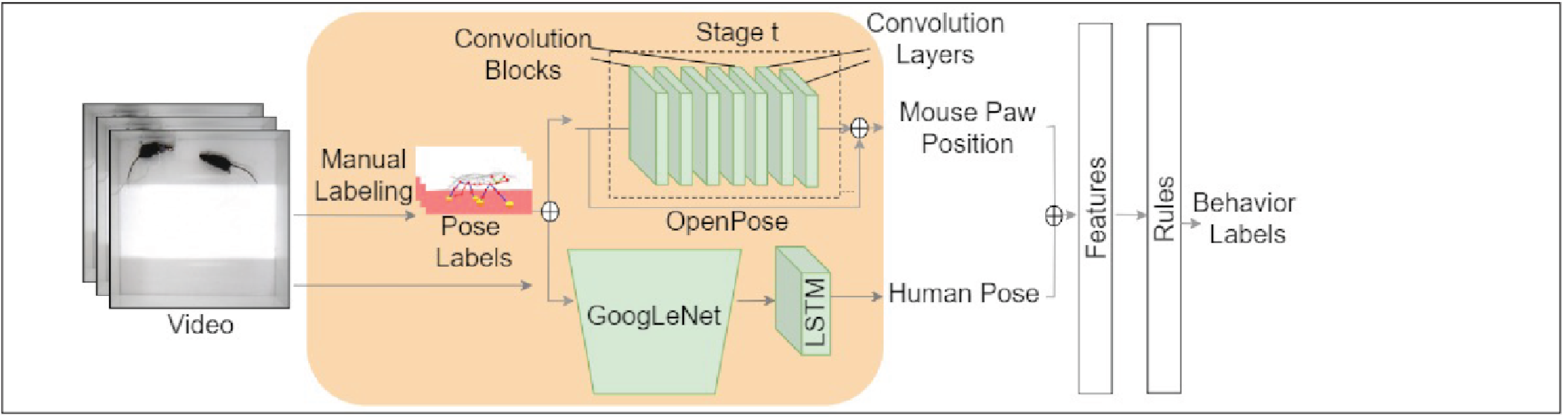}\\
 DeepBehavior~\cite{Arac2019} \\
\includegraphics[width=\textwidth]{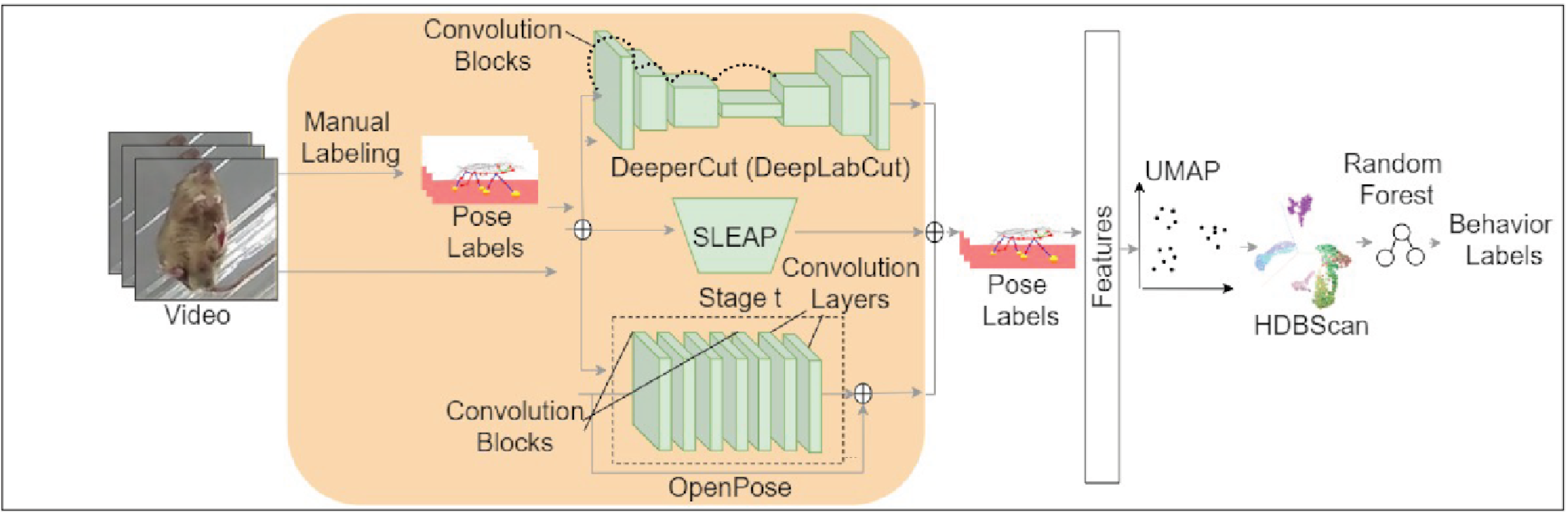}\\
 Behavioral Segmentation of Open Field in DeepLabCut (B-SOiD)~\cite{Hsu2021} \\
\includegraphics[width=\textwidth]{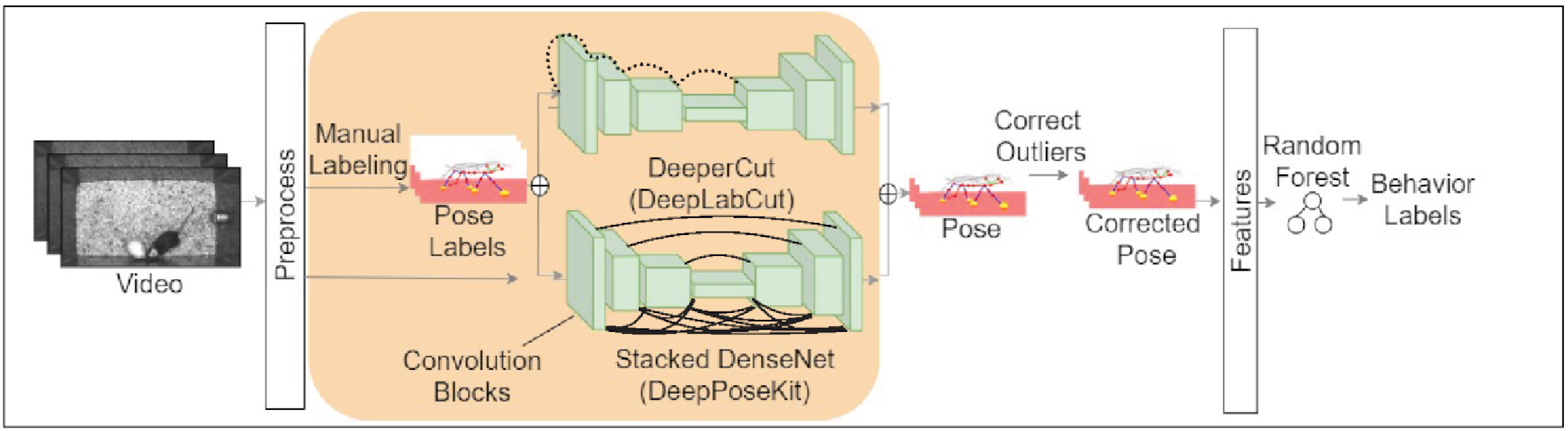}\\
 Simple Behavior Analysis (SimBA)~\cite{Nilsson2020} \\
\includegraphics[width=\textwidth]{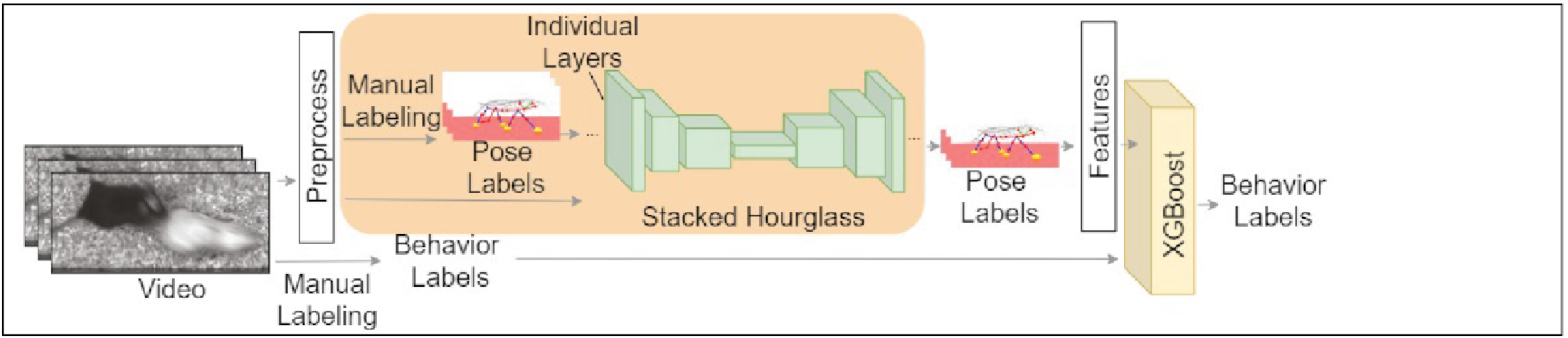}\\
 Mouse Action Recognition System (MARS)~\cite{Segalin2021} \\ 
\end{tabular}
\end{center}
\caption{Pose-based Animal Behavior Classification Frameworks}
\label{fig:posebased}
\end{figure*}

\begin{figure*}
\begin{center}
\setlength\tabcolsep{0.5pt}
\begin{tabular}{ p{0.5\textwidth} p{0.5\textwidth} }
\includegraphics[width=0.5\textwidth]{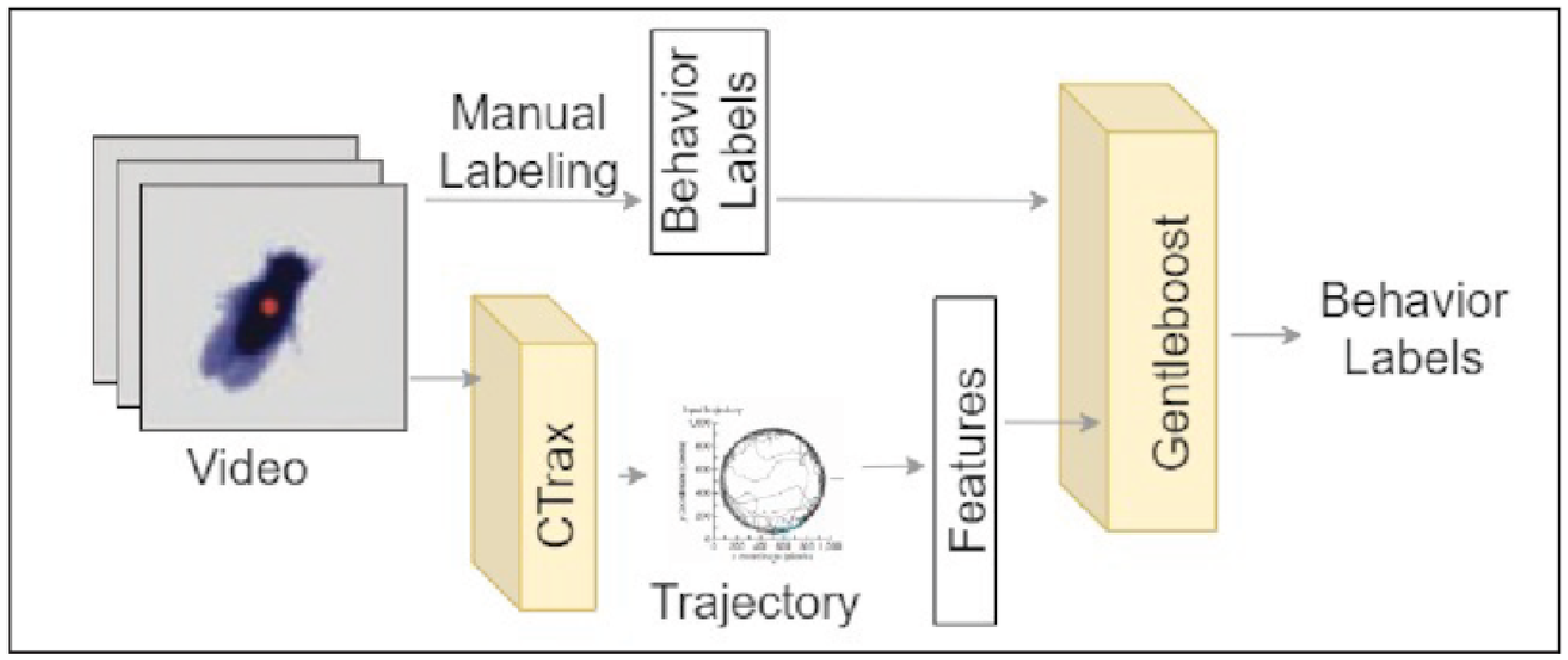} &
\includegraphics[width=0.5\textwidth]{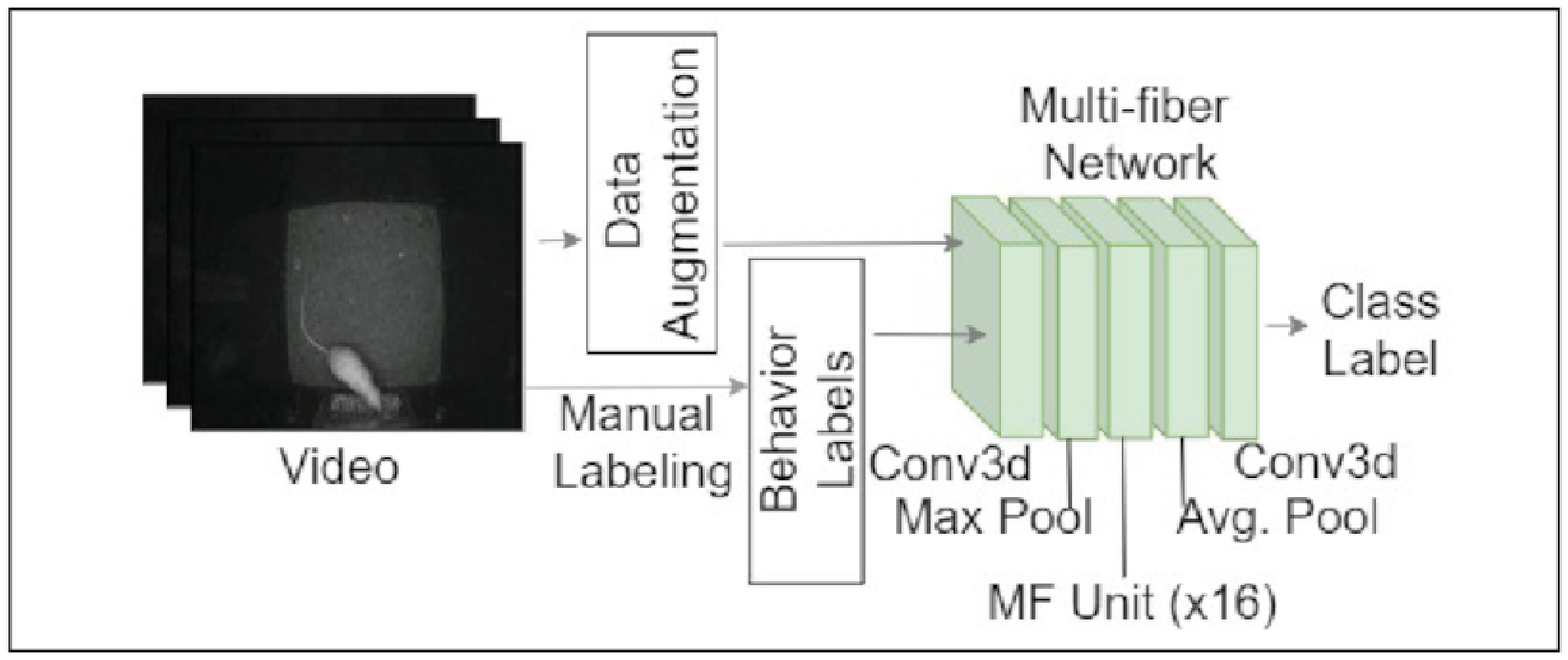} \\
\centering{Janelia Automatic Animal Behavior Annotator (JAABA)~\cite{Kabra2013}} &
\centering{Multi-fiber Network (Van Dam et al., 2020)} \\
\end{tabular}
\begin{tabular}{ c }
\includegraphics[width=\textwidth]{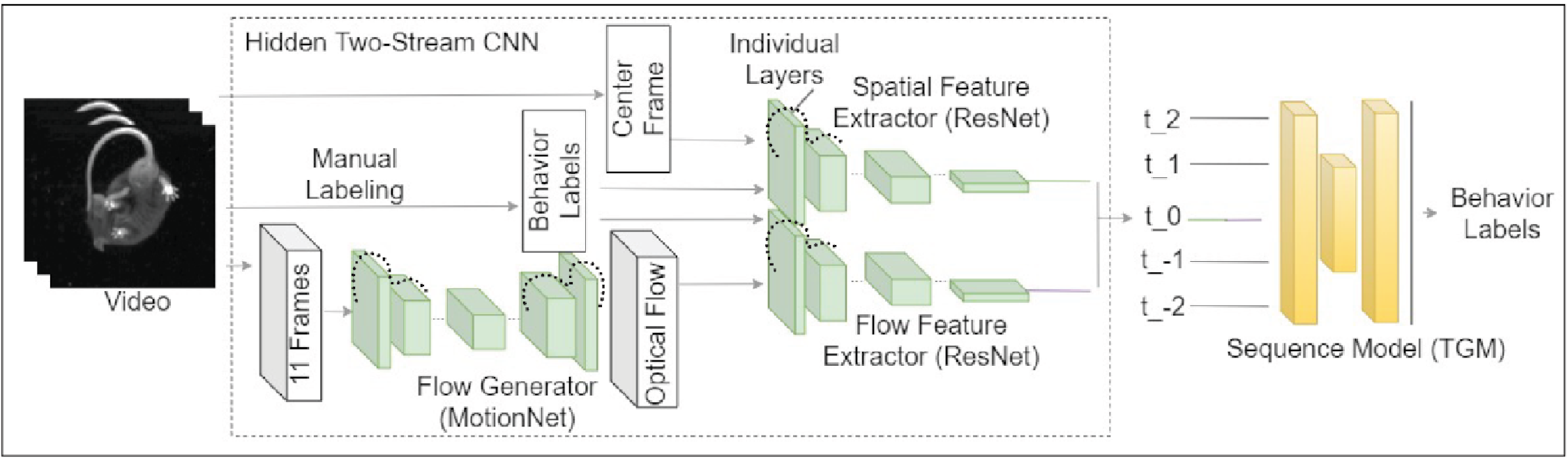} \\
DeepEthogram~\cite{Bohnslav2021} \\
\end{tabular}
\caption{Supervised Animal Behavior Classification Frameworks that Operate on Pixel Values without Pose Estimation}
\label{fig:SNoPE}
\end{center}
\end{figure*}


\begin{figure*}[h]
\begin{center}
\begin{tabular}{ c }
\includegraphics[width=\textwidth]{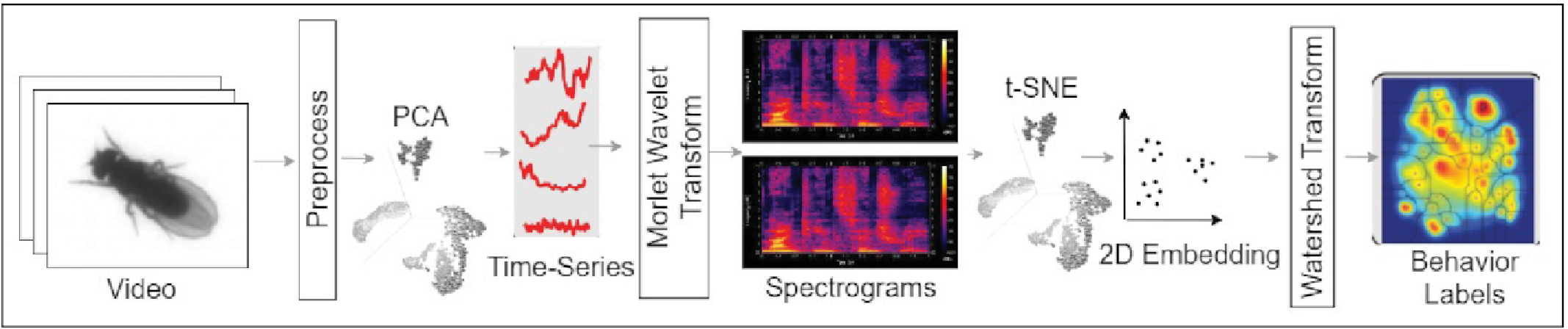} \\
MotionMapper~\cite{Berman2014} \\
\end{tabular}
\setlength\tabcolsep{0.5pt}
\begin{tabular}{ p{0.5\textwidth} p{0.5\textwidth} }
\includegraphics[width=0.5\textwidth]{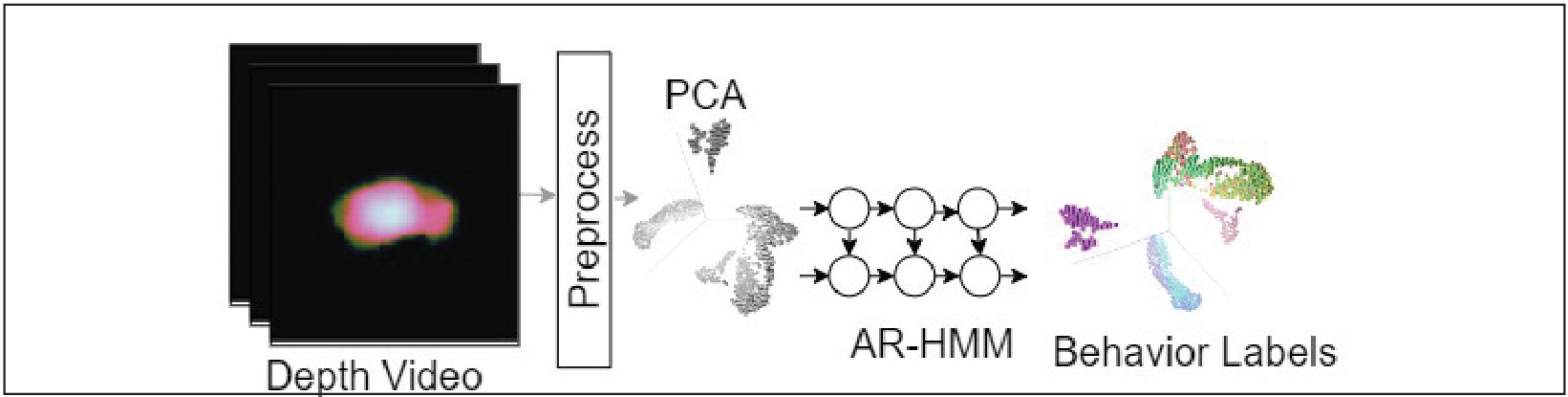} &
\includegraphics[width=0.5\textwidth]{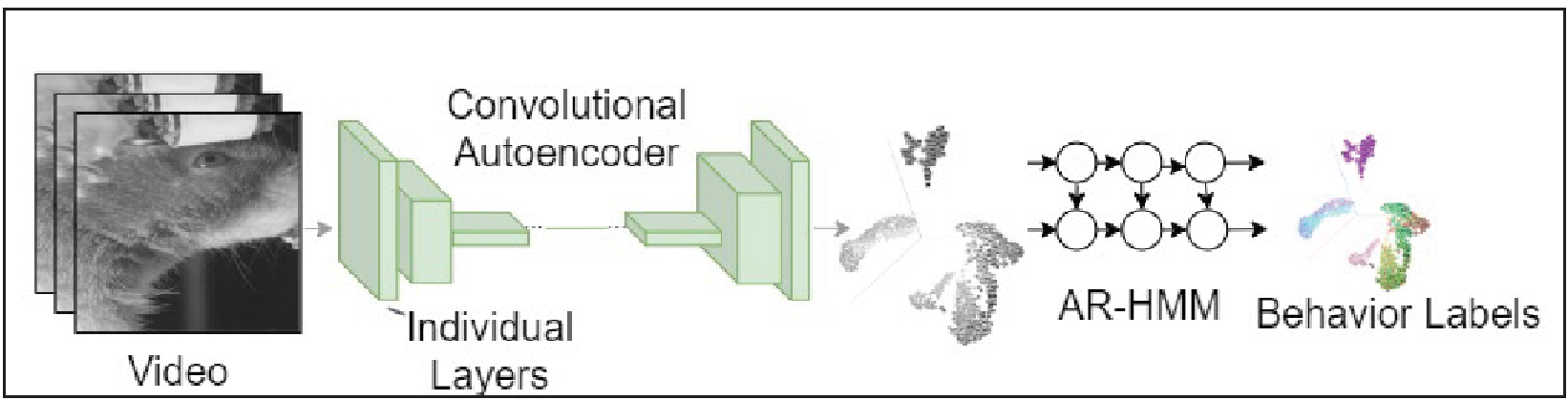} \\
\centering{Motion Sequencing (MoSeq)~\cite{Wiltschko2015,Wiltschko2020}}&
\centering{BehaveNet~\cite{Batty2019}} \\
\end{tabular}
\begin{tabular}{ c }
\includegraphics[width=\textwidth]{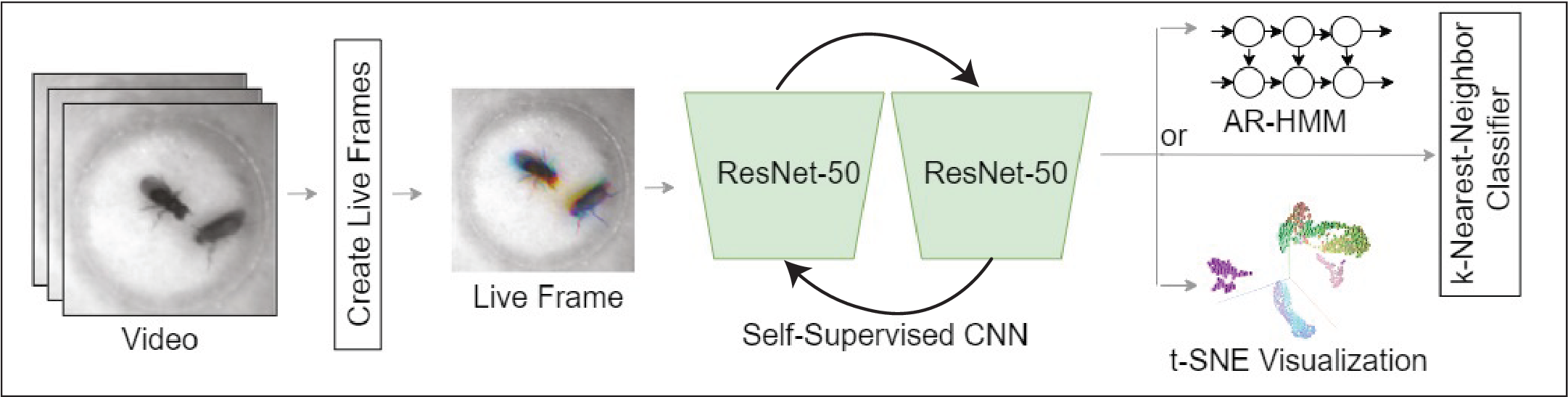} \\
Selfee (Jia et al., 2022) \\
\end{tabular}
\caption{Unsupervised Animal Behavior Classification Frameworks that Operate on Pixel Values without Pose Estimation.} 
\label{fig:USNoPE}
\end{center}
\end{figure*}

\section{Discussion and Future Directions}\label{sec:discussion}

\noindent This section discusses future research that investigates adaptations of action recognition for animal behavior classification.

\vspace{10pt}
\noindent Action recognition has many practical applications as many professions use automated methods to interpret motion from publicly available or privately acquired videos. Although we are motivated by problems with behavior analysis in neuroscience, these research topics are relevant to several application spaces. Recognizing player movements in gaming~\cite{Bloom2012}, monitoring infants~\cite{Moccia2019}, preventative healthcare for the elderly, and hand-object pose estimation in augmented and virtual reality environments~\cite{Doosti2020} are examples.

\vspace{10pt}
\noindent There are opportunities to further evaluate human action recognition methods on animal behavior classification problems. Human action recognition methods are mature compared to emerging developments in animal behavior classification~\cite{JIANG2022103483}. Adapting CNN-based action recognition methods not yet considered in this domain would be beneficial to address domain specific problems like occlusion from hidden limbs on small subjects,  as well as historical challenges like processing high-resolution videos. Action recognition methods work on trimmed videos while many animal behavior methods operate on untrimmed videos with labeled frames. Frames with a broad range of non-class activities may be indicated by a single background class. Accurate evaluation of  new methods on existing benchmark datasets will require special handling when the data varies between these domains. Many CNN architectures operate on low resolution videos. Higher resolution data, or increased kernel sizes (for detecting long range dependencies) increase computation time. There is a need to consider CNN-based systems that are customized to address the speed accuracy trade-off. The following research topics are relevant:

\vspace{10pt}
\noindent\textbf{Fully Unsupervised Behavior Classification:} Behavior analysis is important for diagnosing and treating neurological disorders~\cite{MATHIS20201}. The human action recognition techniques we reviewed were supervised or semi-supervised, requiring labeled videos. Manual labeling of videos is time consuming and inconsistent~\cite{gulinello2019rigor}. Unsupervised animal behavior classification systems~\cite{Berman2014,Wiltschko2015,Wiltschko2020,Batty2019,Jia2022} are few, and open challenges remain. Our taxonomy (Section~\ref{sec:taxonomy}) shows that many methods that are broadly classified as unsupervised are discriminative and pose-based.  Many steps still require hand-crafted features, or labels in complex systems. Unsupervised classification systems would remove manual scoring and inconsistencies that might skew results~\cite{gulinello2019rigor,Bohnslav2021,Segalin2021}. In addition, new micro-scale behaviors that are not pre-defined in a behavior collection may be discovered. Approaches that operate on unstructured data should be able to identify behaviors at different time scales and organize behavior in a hierarchical structure.

\vspace{10pt}
\noindent \textbf{Action Recognition in Crowds and Complex Social Environments:} Many behavioral studies aim to understand complex social behaviors of multiple subjects in intricate environments. In these scenarios, identifying the behavior of multiple subjects performing individual actions in the same frame is challenging. Pose-based action recognition systems that use two-stage inference are more prone to errors when there are multiple overlapping body parts. A common error is the assignment of the same limbs to multiple subjects.  In the domain of animal behavior classification, CS-IGANet and other works~\cite{BurgosArtizzu2012} can classify multi-subject behaviors  in social datasets like the CRIM13 resident-intruder mouse~\cite{BurgosArtizzu2012} dataset. However, these approaches do not predict individual actions of each subject. SIPEC~\cite{Marks2022} is one of the first examples to have success identifying behaviors of individual subjects in a social setting. There is a need to investigate networks that evaluate dense activity (swarms) in \emph{in-the-wild} conditions.

\vspace{10pt}
\noindent \textbf{Robust Multi-modal Action Recognition:} Multimodal data capture has been used to describe an animal's dynamics in social interactions. Multi-modal sensing for behavior recognition is important as a more complete behavior representation may be achieved by combining heterogenous information from different sensory devices. However, it is a challenge to fuse and work with motion  information across different sensory devices, and when there are significant variations in environmental lighting conditions. Networks that operate on multimodal data from different imaging modalities would be effective for interoperability.

\section{Conclusion} \label{sec:conclusion}

Automated classification of activity performed by humans in videos is an important problem in computer vision with applications in many fields. We presented a thorough and detailed review of CNN architectures developed for two related tasks: human action recognition, and human pose estimation. We traced the development of these network architectures, examining design decisions that increased performance on benchmark image and video datasets. We focused our paper on techniques that have been modified for animal behavior classification tasks. Our taxonomy for animal behavior classification frameworks is more comprehensive than others because we consider model inputs and system components that influence more varied levels of supervision. We discussed and compared the components of several of these behavior classification systems, highlighting challenges and future research directions.  We believe this survey will inspire future work in action recognition, its adaptation for animal behavior classification, and new evaluation metrics that compare neural network performance across different levels of supervision.

\bibliographystyle{ACM-Reference-Format}
\bibliography{ms.bib}

\end{document}